\documentclass{article}

\usepackage{PRIMEarxiv}

\usepackage[utf8]{inputenc} % allow utf-8 input
\usepackage[T1]{fontenc}    % use 8-bit T1 fonts
\usepackage{url}            % simple URL typesetting
\usepackage{booktabs}       % professional-quality tables
\usepackage{amsfonts}       % blackboard math symbols
\usepackage{nicefrac}       % compact symbols for 1/2, etc.

\usepackage{pgfplotstable}
\usepgfplotslibrary{groupplots}
\renewcommand{\arraystretch}{1.4} % Adjust row spacing

\usepackage{booktabs} % for professional tables
\usepackage{makecell}
\usepackage{longtable}

\PassOptionsToPackage{table,dvipsnames}{xcolor}

\usepackage[table,dvipsnames]{xcolor} 
\usepackage{colortbl}
\usepackage{subcaption}

\usepackage{placeins}
\usepackage{multirow}
\usepackage{fancybox} 
\usepackage{pgfmath}
\usepackage{amsmath}
\usepackage{amssymb}
\usepackage{mathtools}
\usepackage{amsthm}
 
\usepackage{pgfplots}
\pgfplotsset{compat=1.18}
\usepackage{tikz}
\usepackage{tcolorbox} 
\usepackage{caption}
\usepackage{placeins}
\usepackage{color}
\usepackage{fancybox}
\usepackage{mdframed}
 
\definecolor{bgcolor}{RGB}{245,245,245}
\definecolor{framecolor}{RGB}{200,200,200}
\definecolor{titlecolor}{RGB}{0,102,204}
\definecolor{MistyRose4}{rgb}{0.54, 0.47, 0.47}
\definecolor{LemonChiffon4}{rgb}{0.54, 0.54, 0.47}
\definecolor{NavajoWhite4}{rgb}{0.54, 0.45, 0.34}
\definecolor{Tan4}{rgb}{0.55, 0.37, 0.16}
\definecolor{DarkGoldenrod4}{rgb}{0.55, 0.41, 0.08}
\definecolor{Burlywood4}{rgb}{0.54, 0.43, 0.26}
\definecolor{keywordcolor}{rgb}{0.7, 0.1, 0.1}   % red
\definecolor{tacticcolor}{rgb}{0.0, 0.1, 0.6}    % blue
\definecolor{commentcolor}{rgb}{0.1, 0.5, 0.1}   % grey
\definecolor{symbolcolor}{rgb}{0.0, 0.1, 0.6}    % blue
\definecolor{sortcolor}{rgb}{0.1, 0.5, 0.1}      % green
\definecolor{attributecolor}{rgb}{0.7, 0.1, 0.1} % red
\definecolor{springcolor}{rgb}{0, 0, 0}

\definecolor{orange}{cmyk}{0,0.5,1,0} 
\definecolor{SkyBlue}{rgb}{0.53, 0.81, 0.92}
\definecolor{Periwinkle}{rgb}{0.8, 0.8, 1.0}
\definecolor{Turquoise}{rgb}{0.25, 0.88, 0.82}

\usepackage{listings}

\usepackage{algorithm}
\usepackage{algpseudocode} % algorithmicx 需要这个宏包

\usepackage{authblk}
\usepackage{microtype}      % microtypography
\usepackage{lipsum}
\usepackage{fancyhdr}       % header
\usepackage{graphicx}       % graphics
\graphicspath{{media/}}     % organize your images and other figures under media/ folder

\title{ A Combinatorial Identities Benchmark for Theorem Proving via Automated Theorem Generation
}
\author{\textbf{Beibei Xiong}$^{1}$ \quad 
        \textbf{Hangyu Lv}$^{2}$ \quad 
        \textbf{Haojia Shan}$^{1}$ \quad 
        \textbf{Jianlin Wang}$^{2}$ \quad 
        \textbf{Zhengfeng Yang}$^{1}$\quad 
        \textbf{Lihong Zhi}$^{3}$}

\affil{\normalsize $^{1}$East China Normal University \quad 
        $^{2}$Henan University \quad 
        $^{3}$Chinese Academy of Sciences}

\affil{\normalsize  \texttt{\{bbxiong,hjshan\}@stu.ecnu.edu.cn}, \texttt{hangyvlu@gmail.com}, 
         \texttt{jlwang@henu.edu.cn}, 
        \texttt{zfyang@sei.ecnu.edu.cn}, \texttt{lzhi@mmrc.iss.ac.cn}}

\begin{document}
\maketitle

\begin{abstract}
Large language models (LLMs) have significantly advanced formal theorem proving, yet the scarcity of high-quality training data constrains their capabilities in complex mathematical domains.
Combinatorics, a cornerstone of mathematics, provides essential tools for analyzing discrete structures and solving optimization problems. However, its inherent complexity makes it particularly challenging for automated theorem proving (ATP) for combinatorial identities.  
 To address this, we manually construct  \textit{LeanComb}, combinatorial identities benchmark in Lean, which is, to our knowledge, the first formalized theorem proving benchmark built for combinatorial identities. We develop an \textbf{A}utomated \textbf{T}heorem \textbf{G}enerator for \textbf{C}ombinatorial \textbf{I}dentities, ATG4CI, which combines candidate tactics suggested by a self-improving large language model with a Reinforcement Learning Tree Search approach for tactic prediction.
By utilizing ATG4CI, we generate a \textit{LeanComb}-Enhanced dataset comprising $260$K combinatorial identities theorems, each with a complete formal proof in Lean, and experimental evaluations demonstrate that models trained on this dataset can generate more effective tactics, thereby improving success rates in automated theorem proving for combinatorial identities.
\end{abstract}

% % keywords can be removed
% \keywords{Automated Theorem Generation\and Large Language Models\and Lean Theorem Prover\and Combinational Identities}

\section{Introduction}
The formal theorem proving entails translating theorems and their proofs into a machine-readable format, which allows computers to autonomously execute logical reasoning and verification, thereby ensuring the absolute correctness of the theorems~\cite{geuvers2009proof}.
This process is vital to eliminate potential reasoning errors. Formal proofs have been successfully applied to significant results such as the Kepler Conjecture~\cite{hales2017formal}, the Four Color Theorem~\cite{gonthier2008formal}, and the Clausen-Scholze Conjecture~\cite{theLiquidTensorExperiment}. 
However, interactive theorem proving (ITP) is labor-intensive, particularly when tackling complex theorems, demanding a high degree of expertise from human specialists. For instance, the formalization of the Kepler Conjecture spanned 11 years and involved more than 20 researchers, culminating in a proof through hundreds of thousands of logical steps~\cite{hales2017formal}.

To reduce the high cost of manual formalization, a practical approach is to implement automated theorem proving (ATP), which streamlines the interaction between human experts and proof assistants.
In recent years, with the rapid advancement of artificial intelligence (AI), 
large language models (LLMs) have begun to play a role in ATP, integrating with proof assistants to enhance the capabilities of theorem provers. Some approaches exploit the generative capabilities of LLMs to build complete proofs of theorems in a single decoding process~\cite{jiang2022draft,xin2023lego,cai2024internlm2}. At the same time, other mainstream methods utilize fine-tuned models to generate single-step candidate proof tactics based on the current state of the proof~\cite{wang2023dt,r5,jiang2022thor,polu2020,r18}, gradually building complete proofs by integrating various search algorithms. The works described above primarily focus on fundamental mathematical problems, typically at the high school and undergraduate levels.

Regarding advanced mathematics or more complex mathematical fields, theorem provers based on LLMs still face significant challenges. AlphaProof and AlphaGeometry 2 ~\cite{trinh2024solving} successfully solved four of the six problems in this year's International Mathematical Olympiad (IMO) but have yet to tackle the remaining two combinatorics problems~\cite{lesswrong_2024}. One of the main issues lies in the scarcity of training data, which cannot impact training LLMs. For example, while the Lean formal proof language library, Mathlib4~\cite{MathlibCommunity}, contains more than 180K theorems, the distribution between different mathematical branches is uneven, with only over one hundred in combinatorics.

To break through this predicament, we choose combinatorics as our starting point~\cite{stanley1986enumerative,krantz2010episodic}. Based on Lean~\cite{r8}, a proof assistant widely embraced by mathematicians, we build a combinatorial identities benchmark, \textit{LeanComb}, and introduce an automated theorem generator that combines a self-improving large language model (LLM) ~\cite{huang2022large} with a reinforcement learning (RL) search method inspired by Monte Carlo Tree Search (MCTS) for generating new theorems and their proofs.
To our knowledge, \textit{LeanComb} is the first formalized theorem proving benchmark explicitly created for combinatorial identities.
Fine-tuning and evaluating multiple language models with the \textit{LeanComb} benchmark and \textit{LeanComb}-Enhanced dataset, the experimental results show that the models achieved a proof success rate of up to $26\%$, with each fine-tuned model improving success rates by over $17\%$ compared to the baseline models. The main contributions of this paper are summarized as follows:
\begin{itemize}
    \item
We introduce \textit{LeanComb}, a benchmark for combinatorial identities, comprising $418$ identities and $209$ lemmas in the training set and $100$ identities in the test set.
    \item
We propose a data augmentation method that combines a self-improving language model with an MCTS-inspired search algorithm. Based on the \textit{LeanComb} benchmark, a \textit{LeanComb}-Enhanced dataset containing $260,466$ theorems is automatically generated.  
    \item
Extensive experiments demonstrate that models trained on the \textit{LeanComb}-Enhanced dataset significantly improve the success rate of automated proofs for combinatorial identities. Moreover, evaluations on the miniF2F dataset also show improved performance in automated theorem proving.

\end{itemize}

\section{Related Work}

Automated theorem proving has gained prominence in artificial intelligence~\cite{saxton2019analysing, wang-etal-2023-dt,DBLP:journals/corr/abs-2410-15700}, yet its application remains largely restricted to relatively simple mathematical problems and struggles with proving more complex statements. Data scarcity and its uneven distribution across different mathematical domains is a key challenge.

Among the largest publicly available datasets in AI4Maths, Numina~\cite{li2024numinamath} contains 860K competition-level math problems with solutions following the Chain of Thought (CoT)~\cite{cotgoogle} reasoning paradigm. In contrast, Mathlib~\cite{MathlibCommunity}, the community-maintained Lean formal mathematics dataset, provides a rigorously structured repository spanning algebra, number theory, and combinatorics. However, both datasets rely heavily on manual formalization, leading to limited coverage and uneven representation across specialized mathematical fields.

To address the limitations of data scarcity, researchers have explored automated theorem generation~\cite{piotrowski2018atpboost}. In recent years, theorem generation  techniques based on neural networks and LLMs have emerged as a promising direction, opening up new possibilities for both theorem generation and automated proving ~\cite{gauthier2019deep,r1}.
Existing approaches leverage various strategies for theorem generation: INT~\cite{r1} integrates inequalities to construct new theorems, Leandojo~\cite{r5} retrieves and synthesizes formal statements from Mathlib, and MetaGen~\cite{MetaGen2020} transplants proof trees while employing reinforcement learning to align theorem generation with human reasoning. Meanwhile, another class of generators relies on LLMs to assist in theorem generation. Specifically, these models are often employed as premise selectors to identify key reasoning steps or to iteratively sample from existing theorems, generating new ones that adhere to specific rules~\cite{urban2020first, ds-prover, palermo2022synthetic, polu2020, lime}. Furthermore, the PACT method~\cite{r18} extracts data from theorems and generates nine distinct language modeling tasks, enabling data augmentation within the Lean theorem prover.

However, most existing theorem generators expand datasets by extracting sub-theorems from complete theorems or synthesizing new ones. Against this backdrop, we combine
LLMs’ broad knowledge coverage with RL’s exploration-driven capabilities to conduct tactic prediction research based on our manually constructed benchmark, \textit{LeanComb}.
Our goal is to generate high-quality, novel theorems that contribute to developing data resources in specialized mathematical fields while advancing the capabilities of automated theorem-proving systems.

\section{A Combinatorial Identities Benchmark}
 In this section, we introduce \textit{LeanComb}, a manually formalized benchmark for combinatorial identities based on Lean 4~\cite{r8}. To the best of our knowledge, \textit{LeanComb} is the first formal theorem benchmark specifically dedicated to combinatorial identities, covering a broad range of combinatorics topics, including classical and modern identities. The dataset aims to provide a rigorously verified and formalized collection of combinatorial identities that can be used to evaluate the performance of ATP tools. Lean is an interactive theorem prover and programming language developed by Microsoft Research, which is based on dependent type theory~\cite{chlipala2022certified}. It is supported by a comprehensive and actively maintained mathematical library, Mathlib4~\cite{MathlibCommunity}, developed by its community. In the following paper, Lean refers to Lean 4 by default.

Our dataset is mainly derived from the classical combinatorial mathematics literature~\cite{spivey2019art,gould1972combinatorial,shi2001combinatorial}.
We manually translate informal statements into formal definitions and theorems, including their statements and proofs. The combinatorial identities in the training set of {\it LeanComb} are primarily selected from \cite{spivey2019art}, while those in the test set are drawn from \cite{gould1972combinatorial,shi2001combinatorial}. 
 The benchmark consists of $727$ combinatorial identities, with $627$ theorems in the training set ($418$ identities and $209$ lemmas) and $100$ in the test set (100 identities).

 \begin{table}[htp!]
\centering
\scalebox{1}{ % 缩放比例调整为 85%
\begin{tabular}
{
|@{\hbox to .1em{\hss}}l@{\hbox to .1em{\hss}}
|@{\hbox to .1em{\hss}}l@{\hbox to .1em{\hss}}
|c|c
%|@{\hbox to .3em{\hss}}c@{\hbox to .3em{\hss}}
%|@{\hbox to .3em{\hss}}c@{\hbox to .3em{\hss}}
|}
%{|l|l|c|c|}
\hline
\multicolumn{2}{|c|}{ Theorem Categories}  &  Training Set  &   Test Set  \\ \hline
\multirow{3}{*}{ Basic Techniques}  & The Generalized Binomial Coefficient & 42 & 16 \\ \cline{2-4} 
& Absorption Identity & 35 & 12 \\ \cline{2-4} 
& Binomial Inversion & 46 & 6 \\ \hline
\multirow{3}{*}{ Combinatorics}  & Choosing with Replacement & 26 & 6 \\ \cline{2-4} 
&  Alternating Binomial Sums \& Involutions & 23 & 6 \\ \cline{2-4} 
&  Inclusion-Exclusion & 27 & 5 \\ \hline
\multirow{3}{*}{ Calculus}  & Differentiation & 20 & 5 \\ \cline{2-4} 
& Integration & 26 & 3 \\ \cline{2-4} 
& Beta Integral \& Gamma Function & 22 & 6 \\ \hline
\multirow{2}{*}{ Probability} &  Binomial \&   Hypergeometric Distributions
 & 13 & 5 \\ \cline{2-4} 
&  Expected Values \&  Moments & 16 & 3 \\ \hline
\multicolumn{2}{|l|}{ Special Numbers}  & 25 & 10 \\ \hline
\multicolumn{2}{|l|}{ Generating Functions}  & 29 & 5 \\ \hline
\multicolumn{2}{|l|}{ Recurrence Relations 
\& Finite Differences}  & 56 & 6 \\ \hline
\multicolumn{2}{|l|}{ Miscellaneous Techniques \& Mechanical Summation }  & 12 & 6 \\ \hline
\multicolumn{2}{|l|}{ Total}  & 418 & 100 \\ \hline
\end{tabular}
}
\vspace{0.6em}
\caption{Theorem categories with training and test set counts}
\label{tab:tb1}
\end{table}

We have formalized $9$ new definitions, encompassing theorems related to Stirling numbers of the first and second kind and their properties. We treat each tactic as a single proof step, e.g., a lemma introduced by the ``$have$'' tactic counts as one step. The maximum number of proof steps in the training set is $98$, with an average of $8.9$ steps. \textit{LeanComb} Benchmark contains many classical theorems, such as:
\begin{align*}
&\text{Vandermonde ' s\ Identity}:{\textstyle \sum_{k=0}^{r}} C_{n}^{k}C_{m}^{r-k}=C_{n+m}^{r},\\
&
\text{Cassini's Identity}: \forall n \in \mathbb{Z}, \quad F_{n+1} F_{n-1} - F_n^2 = (-1)^n,\\
&\text{Parallel\ Summation}: {\textstyle \sum_{k=0}^{m}} C_{n+k}^{k}=C_{n+m+1}^{m}.
\end{align*}

As shown in Table~\ref{tab:tb1}, all identities in the training set and test sets are classified into $8$ categories, including Combinatorics, Calculus, Probability, and others. The Basic Techniques category contains the most significant samples in training and test sets, with $123$ and $40$ samples, respectively. 
In the “Basic Techniques” category, the Binomial Coefficient and Binomial Inversion, along with the Recurrence Relations \& Differences categories, have relatively higher sample counts, totaling $42$, $46$, and $56$, respectively.

We now use a simple example to demonstrate the formalization of a combinatorial identity in Lean 4. 

\noindent

\begin{equation}\label{e1}
\sum_{k=1}^{n} nC^{k-1}_{n-1} = n\sum_{l=0}^{n-1}C^{l}_{n-1}.
\end{equation}

\begin{lstlisting}[caption={A combinatorial identity in Lean 4.}, breaklines=true]
theorem sum_mul_congr {n : ℕ}:
∑ k in Ico 1 (n + 1), n * choose (n-1) (k-1) = n * ∑ l in range n, choose (n-1) l := by
    rw [mul_sum]
    rw [sum_Ico_eq_sum_range]
    simp
\end{lstlisting} \label{l1}
 
We formalize the statement of the combinatorial identity~(\ref{e1}), named $sum\_mul\_congr$, in Line 2, with the detailed proof process presented in Lines 3 through 5. In particular, Line 3 applies the ``$rw \ [mul\_sum]$'' tactic to factor the constant $n$ out of the summation. Then, in Line 4, the ``$rw \ [sum\_Ico\_eq\_sum\_range]$'' tactic is used to perform variable substitution. Finally, the ``$simp$'' tactic automatically simplifies the subgoal, successfully advancing the proof to the ``$no \ goals$'' state.
The tactics employed must be predefined or established within the Lean library or associated files. Since each theorem depends on these tactics, discovering additional high-quality tactics is essential for 
advancing the generation of automated theorems. 

 %\vspace{-.3em}
\begin{figure*}[htp!]
\centering
\includegraphics[width=0.9\textwidth]{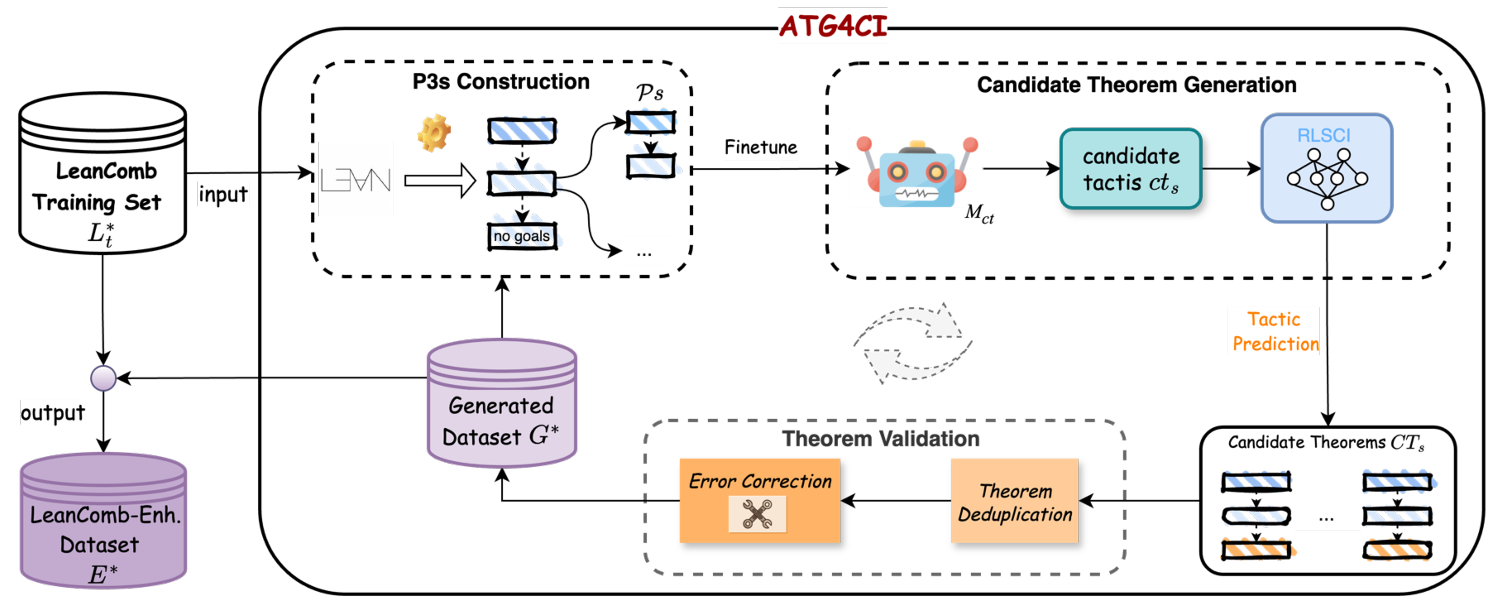} % Reduce the figure size so that it is slightly narrower than the column.
\caption{The Framework of ATG4CI.}
\label{fig2}
\end{figure*}

\section{Lean4Kit}

Lean4Kit, built on Lean 4, provides offline data extraction and interaction capabilities, significantly enhancing automated theorem generation and proving. First, Lean4Kit extracts state-tactic pairs from Lean code to construct a training set. Second, it enables programmatic interaction with the Lean system, supporting tactic prediction and automated theorem proving.
% More details will be found in Appendix~\ref{AppendixB}.

\noindent
\textbf{Data Extraction.} 
Lean4Kit can access the current proof state in run-time during the Lean theorem proving process. By interacting with the Lean system, it tracks the proof process and dynamically updates the proof state.

\textit{State Information}: Our toolkit constructs the sequence of proof states by extracting the Lean code and the proof goal. 
Additionally, it captures the error message if an error occurs after invoking a tactic and identifies whether a proof state has been successfully resolved after executing a tactic.

\textit{Premises and Tactics}: 
For the Lean code of a given theorem, the toolkit extracts all premises and variable types, ensuring that the premises' names align with those of the Lean libraries. If the proof state is valid at each step of the proof, it also captures the corresponding tactic used.

\noindent
\textbf{Interaction.} Another key feature of the toolkit is its ability to interact with Lean 4. It can initialize the proof state, update it after executing the tactic, and trace the feedback information, including error messages or proof completion. 
Lean4kit can also be utilized for tactic prediction, filtering candidate tactics that can efficiently modify the proof state. Below is a detailed description of the interactive features:

\textit{get\_init\_state (theorem)}: Given a theorem, our toolkit obtains its initial proof state and extracts data from that state. If a theorem, including statement and proof, fails validation during interaction, its initial state will be marked as an error.

\textit{run\_tactic (state, tactic)}: After applying a tactic to a given state,  the updated proof state is returned. If an inapplicable tactic is executed, an error state and the corresponding error message are generated. If the input state is already an error, the result will also be an error.

\section{The Framework of Automated Theorem Generator for Combinatorial Identities}

This section introduces ATG4CI, an iterative framework for automated theorem generation for combinatorial identities. We propose a tactic prediction method combining a fine-tuned LLM with a reinforcement learning algorithm. The framework generates candidate theorems using tactic prediction, verifies them through theorem validation, and compiles the results to build an enhanced dataset.
The generated theorem dataset can fine-tune the LLM through iterative refinements, enhancing its tactic prediction capabilities. Meanwhile, with its enhanced predictive capability, the improved LLM can boost the quantity and quality of automatically generated theorems.
  
As shown in Fig.~\ref{fig2}, ATG4CI comprises three essential stages: \textit{Partial  Proof Paths (P3s) Construction}, \textit{Candidate Theorem Generation}, and \textit{Theorem Validation}. The procedure begins with the $\textit{LeanComb}$ training set $L_t^*$, consisting of formalized combinatorial identities, serving as the foundational data for data augmentation and model fine-tuning. Building upon this, the pipeline enters the \textit{P3s Construction} phase, where the Lean4Kit tool is employed to transform each theorem into a proof tree. Their corresponding state-policy pairs are extracted from these trees. The states are then corrected to ensure their quality and correctness. 

Following this, in the \textit{Candidate Theorem Generation} phase (top right of Fig. \ref{fig2}), the procedure begins with generating candidate tactics using a fine-tuned model $M_{ct}$. The candidate tactics are then refined by selecting the most appropriate ones for each partial proof path (\textit{P3}) through a Reinforcement Learning-based search tailored to address the specific requirements of the combinatorial identities domain. This tactic prediction process is repeated to ultimately derive candidate theorems,  followed by the \textit{Theorem Validation} stage (bottom of Fig. \ref{fig2}), where redundant theorems are eliminated, and the correctness of the theorems is verified. Specifically, this stage consists of two key steps: \textit{Theorem Deduplication} and \textit{Error Correction}, both of which ensure the uniqueness and correctness of the dataset. 

Ultimately, the validated theorems are compiled into a new dataset $G^*$, fed to train the model $M_{ct}$, enhancing its ability to generate more effective candidate tactics.  Finally, the generated dataset is combined with the \textit{LeanComb} training set $L_t^*$ to form the \textit{LeanComb}-Enhanced dataset $E^*$, which subsequently improves the performance of automated theorem proving. 
%We demonstrate the procedure of ATG4CI using Example 1 in Listing 1, a typical example is also provided in Appendix C for reference.

\subsection{Partial Proof Paths (\textit{P3s}) Construction}
 
We focus on the \textit{P3s} construction process, explaining how to use Lean4Kit to extract partial proof path  $\mathcal{P}_s$ from formalized theorems. To facilitate the exploration of new proof paths for \textit{P3s}, we visualize all tactics within the proof environment, capturing the state transitions that occur before and after applying each tactic. A fully formalized proven theorem, consisting of $n$ tactics, can be represented as a proof tree: the root theorem forms the root node, tactics are the edges, intermediate states are the child nodes, and the ``$no\ goals$'' state is the leaf node. From this tree, $\mathcal{P}_s = \{p_{i}\}_{i=1}^{n-1}$ is extracted by tracing the paths from the root to the intermediate states.

As shown in Example 1, the root theorem, represented by the root node in (\ref{e1}), transits through three sequential tactics and eventually reaches the “no goals” state at the leaf node,  resulting in a four-layer proof tree. From this tree, two \textit{P3s} are extracted: one from the root to the state after applying ``$rw [mul\_sum]$'', and another from the root to the state after applying ``$rw [sum\_Ico\_eq\_sum\_range]$''.

\subsection{Candidate Theorem Generation}

The section explains how to generate candidate theorems from a given partial proof path, which consists of two key steps: candidate tactic generation, produced by a fine-tuned model, and tactic prediction, based on a reinforcement learning search for combinatorial identities (RLSCI).

\noindent
\textbf{Candidate Tactic Generation.}
We start with employing a fine-tuned model $M_{ct}$, trained by the training set of {\it LeanComb}, to generate candidate tactics $ ct_s $ for \textit{P3s}. To enhance the quality and diversity of candidate tactics, we adopt an iterative refinement tactic inspired by self-improvement techniques for fine-tuning models. Initially, the model is fine-tuned on the training set of the \textit{LeanComb} benchmark $L_t^*$ and Lean's foundational library, Mathlib4. Given a partial proof path, the improved model can generate $t$ candidate tactics, where $t$ is a prior positive integer.  The model will be continuously refined in subsequent iterations through fine-tuning with the augmented theorems generated from the previous iteration. This iterative framework not only enhances the model's capacity to propose effective tactics but also broadens its exploration of diverse tactic spaces.

\textbf{Tactic Prediction.} The tactic prediction, based on RLSCI, consists of three primary steps: selecting candidate tactics, expanding the \textit{P3s}, and back-propagating values from the leaf nodes to the root. These steps are iteratively performed until the proof is completed or no viable tactic is identified. If successful, the process discovers a complete proof path for the root theorem \(RT\); otherwise, it generates a candidate proof path \(cp_k\) for \(k = 0, 1, \ldots, s\).  

Our RL framework comprises a critic model \(C_{\theta}\) and a policy model \(P_{\theta}\). Completed proof nodes are assigned a value of $1$, while failed nodes are assigned a value of  $-1$. Unresolved nodes are evaluated using the Polynomial Upper Confidence Trees (PUCT) method~\cite{silver2017mastering}:  
%\vspace*{-0.3em}
\begin{equation*}
Q_{PUCT}(s) = Q(s, t) + c_{puct} \cdot P(s, t) \cdot \frac{\sqrt{ \sum_{b}{N(s,b)}}}{N(s,t)+1},
\end{equation*}
where \(Q(s,t)\) is the estimated value of the state-tactic pair, obtained from the value network or learned from past simulations, and $P(s,t)$ is the probability of selecting a tactic in state \(s\) based on the policy network, and \(c_{puct}\)  is the exploration coefficient. In contrast, \(N(s,t)\) is the count of executions of tactic \(t\) in state \(s\) during the prediction process.  
All successful tactics are stored as training data for LLMs. During backpropagation, values from the leaf nodes are propagated to the root, updating the visit counts \(N(s,t)\) and cumulative action values. 

The process aims to propagate proof paths as far as possible, terminating when the proof is complete or when no viable tactic can be found. In the former case, the leaf node is marked as “no goals”, indicating the discovery of a new proof for the root theorem \(RT\). In the latter case, the goals corresponding to the leaf nodes of the candidate proof paths are treated as candidate theorems, \(CT_m = \{CT_i\}_{i=0}^m\).  

To generate new proofs, we incorporate the original root theorem \(RT\) into the hypotheses and then apply tactics from the candidate path until the goal aligns with the target. Once aligned, the “assumption” tactic resolves the goal, completing the proof.
The following example illustrates the above procedure for \textit{ATG4CI}.

\noindent
\textbf{Example 1 Continued.} Below is one of the generated candidate proof paths: 
\begin{lstlisting}
theorem cp_1 {n : ℕ} :
 ∑ k in Ico 1 (n + 1), n * choose (n - 1) (k - 1) = n * ∑ k in range n, choose (n - 1) k := by
    rw [mul_sum]
    -- Prediction start!
    rw [range_eq_Ico]
    rw [sum_Ico_eq_sum_range]
    rw [add_tsub_cancel_right]
    rw [range_eq_Ico]
    -- Prediction end
\end{lstlisting} 
The \textit{P3s} spanning Lines 1 to 3 is extracted from Example 1 during the \textit{P3s} construction phase. It originates from the root node and extends to the child node after applying the tactic \( rw[mul\_sum] \). The tactic prediction process begins at Line 5, where the LLM is prompted to provide $8$ candidate tactics for the current state. Among these, RLSCI successfully selects four new nodes during the first round. The process proceeds with iterative expansion and backpropagation, repeating these steps for each new node and ultimately generating over $100$ candidate proof paths. In one instance, after four steps (Lines 5–8), the prediction process identifies that no feasible tactic exists for a specific path, leaving the subgoal corresponding to its leaf node as follows:
  \begin{equation}
    {\textstyle \sum_{k=0}^{n}} nC_{n-1}^{1+k-1} ={\textstyle \sum_{l=0}^{n}} nC_{n-1}^{l} . 
    \label{e3}
 \end{equation}
 
The subgoal presented in  (\ref{e3}) is treated as a new candidate theorem:

\begin{lstlisting}
theorem CT_1 (n : ℕ) 
  (h : ∑ k in Ico 1 (n + 1), n * choose (n - 1) (k - 1) = n * ∑ l in range n, choose (n - 1) l) : 
  -- Candidate theorem 
  ∑ k in range n, n * Nat.choose (n - 1) k = ∑ x in Ico 0 n, n * Nat.choose (n - 1) x := by
    rw [mul_sum] at h
    rw [range_eq_Ico] at h
    rw [sum_Ico_eq_sum_range] at h
    rw [add_tsub_cancel_right] at h
    assumption
\end{lstlisting}

Line 2 shows that the root theorem is adopted as the new hypothesis \( h \). Then, in Lines 5–8, all tactics along the candidate path are applied sequentially to rewrite \( h \) until it aligns with the target goal. Finally, the tactic “assumption” in Line 9 is used to conclude the proof.

% %\vspace{0.3em}

\subsection{Theorem Validation}

Note that not all candidate theorems are correct or unique, so a validation process, including theorem deduplication and correction, is necessary to retain the valid theorems.

\textbf{Theorem Deduplication}: Candidate theorems are deduplicated from two perspectives. First, textual duplication is identified by comparing the goals, premises, and proof steps. Identical theorems are merged, retaining only one. Second, mathematical equivalence is checked by simplifying redundant terms (e.g., \(+0\), \(-0\), \(*1\), and \(/1\)), ensuring that only one mathematically equivalent version is kept.

\textbf{Theorem Correction}: After deduplication,  as some candidate theorems may still contain expression errors or fail to pass the proof process, correctness refinement is performed. After candidate theorems are verified successfully by interacting with Lean, they are directly added to the generated dataset \(G^*\). Those that fail verification are categorized by error type and corrected using the corresponding correction methods. The corrected theorems are then added to \(G^*\). The following section elaborates on the error types and their corresponding correction tactics.
%\vspace{-.7em}
\begin{itemize}
\item \textit{Incomplete Errors.} An incomplete error occurs when a candidate tactic generates multiple subgoals, all of which are correct, but at least one subgoal remains unproven, preventing the completion of the theorem's proof.
The standard MCTS~\cite{coulom2006efficient} method can be applied to recover incomplete theorems and generate additional proof steps to complete the proof.
% The process involves interacting with Lean through Lean4Kit to verify whether the theorem has been fully proven. If the proof is incomplete, the proof state is restored, and the corresponding node is linked to the MCTS framework to facilitate automated theorem proving

\item \textit{Type Errors.} Type errors arise due to Lean’s representation methods, where variable types in subgoals are not always explicitly stated during interactions with Lean or during the data extraction process. 
% This type of error often results in difficulties extracting type information comprehensively.
For example, if the original theorem's goal contains the term “$(-1:R)$”, the extraction may capture only “$(-1)\uparrow$”. To address this, we identify the type information from the original theorem and annotate the newly generated theorems with the correct type labels.

\item \textit{Logical Errors.} Logical errors occur when applying a candidate tactic generates multiple subgoals, but at least one contains a logical inconsistency, preventing further progress in the proof. For instance, consider the following theorem where \( n \) must be a natural number:
 $\sum_{k=1}^n nC_{n-1}^{k-1} = n\sum_{l = 1}^{n-1}C_{n-1}^l$ .
After applying the tactic “$ rw [sum\_Ico\_succ\_top] $”, two subgoals are generated, one of which is \( 1 \leq n \), which contradicts the definition of natural numbers. Theorems with such logical inconsistencies are considered irreparable at this stage and must be discarded.
\end{itemize}

Then, we briefly introduce the main steps of ATG4CI implemented in Algorithm \ref{alg:ATG4CI}. The procedure takes as inputs the training set of \textit{LeanComb} Benchmark $L^*_t$, the model $M_{ct}$ that provides candidate tactics $ct_s$, search method RLSCI, the maximum number of iterations $n$, and returns the \textit{LeanComb}-Enhanced dataset $ E^{*}$. 

\begin{algorithm}[htp!]
    \caption{The Framework of ATG4CI}
    \label{alg:ATG4CI}  
    \begin{algorithmic}[1]
        \Require The training set of \textit{LeanComb} Benchmark $L^*_t$, the model $M_{ct}$ that provides candidate tactics $ct_s$, search method RLSCI, maximum number of iterations $n$
        \Ensure \textit{LeanComb}-Enhanced dataset $\textbf{E}^{*}$       
        \State $\mathcal{P}_s \gets Construct\_\textit{P3s}(L^*_t)$
        \State $G_{0} \gets L^*_t$
        \For{$i=0 \to n$ }
            \State $M_{ct}^{i} \gets Fintune(G_{i})$
            \State $ct_s \gets Produce\_{ct_s}(\mathcal{P}_s, M_{ct}^{i})$
            \State $G_{i} \gets Generate\_CT_s(ct_s,RLSCI)$
            \State $G_{i}^{*} \gets Validation(G_{i})$
            \State $E^{*} \gets E^{*} + G_{i}^{*}$
        \EndFor\\
        \Return $\textbf{E}^{*}$
    \end{algorithmic}
\end{algorithm}

%\vspace{-0.2em}
% \begin{algorithm}[htp!]
%     \caption{The Framework of ATG4CI}
%     \label{alg:ATG4CI}    \renewcommand{\algorithmicrequire}{\textbf{Input:}}
%     \renewcommand{\algorithmicensure}{\textbf{Output:}}
%     \begin{algorithmic}[1]
%         \REQUIRE   the training set of \textit{LeanComb} Benchmark $L^*_t$, the model $M_{ct}$ that provides candidate tactics $ct_s$, search method RLSCI, maximum number of iterations $n$
%         %%input
%         \ENSURE \textit{LeanComb}-Enhanced dataset $\textbf{E}^{*}$       \STATE  $\mathcal{P}_s \gets Construct\_\textit{P3s}(L^*_t)$
%        \STATE $G_{0} \gets L^*_t$
%         % \STATE  $i \gets 0$
%         \FOR{$i=0 \to n$ }
          
%             \STATE  $M_{ct}^{i} \gets Fintune(G_{i})$
%             \STATE $ct_s \gets Produce\_{ct_s}(\mathcal{P}_s, M_{ct}^{i})$
%             \STATE  $G_{i} \gets Generate\_CT_s(ct_s,RLSCI)$
%             \STATE  $G_{i}^{*} \gets Validation(G_{i})$
%             \STATE  $E^{*} \gets E^{*} + G_{i}^{*}$
%         \ENDFOR
%      \RETURN $\textbf{E}^{*}$
% \end{algorithmic}
% \end{algorithm}
%\vspace{-0.2em}

We first construct partial proof paths  $\mathcal{P}_s$ and initialize the generated dataset as $ G_0 $  based on the \textit{LeanComb} training set $L^*_t$. Subsequently, the following steps are iteratively performed within a predefined maximum number of iterations. First, the model $ M_{ct} $ is fine-tuned using the current generated dataset (as described in Line 4). Next, based on \textit{P3s}, $ M_{ct} $ is employed to generate candidate tactics (line 5). Subsequently, the search algorithm RLSCI is applied to generate candidate theorems (Line 6). The generated candidate theorems must undergo rigorous validation before being incorporated into the generated dataset $ G_{i}^* $ and further integrated into the enhanced dataset $ E^{*}$ (Line 7).

\section{Experiments}

In this section, we employ ATG4CI to construct the \textit{LeanComb}-Enhanced dataset for combinatorial identities.
Subsequently, we evaluate the performance of fine-tuned LLMs in automated theorem proving on both the \textit{LeanComb} test set and the miniF2F dataset.
Our models, trained on both the \textit{LeanComb} benchmark and the Enhanced dataset, consistently achieve higher proof success rates compared to the baseline, demonstrating the effectiveness of our dataset.
% Further details can be found in Appendix \ref{AppendixF}.

\subsection{\textit{LeanComb}-Enhanced Dataset Generated by ATG4CI}

In this experiment, we iteratively fine-tune the general-purpose LLM Llama3.1-8b~\cite{biderman2023pythia} to assist the theorem generation process by providing candidate tactics for tactic prediction. The number of tactics suggested by the model for each node is limited to $16$.

\noindent
\textbf{Tactic Prediction Model Architecture.} 
In tactic prediction process, the RL component of the RLSCI search algorithm is implemented using two neural networks: a policy network and a critic network. Each network comprises two linear layers, each with a dimensionality of $16$. The training process spans a total of $10$ iterations. During each decision-making step, the algorithm performs $100$ simulations, with the total number of full events per iteration fixed at $20$. Furthermore, each node is allowed to propose up to $16$ candidate theorems. The search time is restricted to $300$ seconds.

\begin{table}[htp!]
\centering
\caption{The performance with varying candidate tactic counts.}
\vspace{0.6em} 

\setlength{\tabcolsep}{10pt} % Reduce column padding to fit within width
\renewcommand{\arraystretch}{1.4} % Reduce row height slightly
\begin{tabular}{@{}c|c|c|c|c@{}}
\toprule
\multirow{2}{*}{\textbf{\# Iteration}} 
& \multirow{2}{*}{\textbf{Theorem Types}} 
& \multicolumn{3}{c}{\textbf{\# Candidate Tactics}} \\ 
\cmidrule {3-5} 
& & \textbf{4} & \textbf{8} & \textbf{16} \\

\midrule
\multirow{5}{*}{\textbf{1st Iter}}
& \# Candidate  & 211,087 & 364,789 & 592,811   \\ 
& \# Deduplicated   & 73,385 & 229,541 & 392,818  \\ 
& \# Correct   & 16,296 & 52,220   & 68,771  \\ 
& \# Corrected   & 6,395 & 21,916   & 31,306  \\ 
\hline
\rowcolor{cyan!10}
\multicolumn{2}{c|}{\textbf{Subtotal}} 
& \textbf{22,691} 
& \textbf{74,136} 
& \textbf{100,077} 
\\ 

\midrule
\multirow{5}{*}{\textbf{2nd Iter}} 
& \# Candidate & 247,996 & 376,158 & 440,199   \\ 
& \# Deduplicated   & 185,850 & 225,797 & 238,425  \\ 
& \# Correct   & 69,765 & 132,791   & 135,796  \\ 
& \# Corrected   & 13,912 & 24,147   & 24,593  \\ 
\hline
\rowcolor{cyan!20}
\multicolumn{2}{c|}{\textbf{Subtotal}} 
& \textbf{83,677}  
& \textbf{156,938} 
& \textbf{160,389} 
\\ 

\midrule
\multicolumn{2}{c|}{\textbf{Total}} 
& \textbf{106,368} 
& \textbf{231,074} 
& \textbf{260,466} 
\\ 
  
\bottomrule
\end{tabular}
\label{tb:2}
\end{table}

\noindent
\textbf{Results.}
Table \ref{tb:2} presents the theorems generated by our ATG4CI under different candidate tactic settings ($4$, $8$, and $16$) based on the \textit{LeanComb} training set. The experiment consists of two iterations. In the first iteration, with $16$ candidate tactics, $592,811$ theorems are generated, which are reduced to $392,818$ deduplicated theorems after deduplication. Among these, $68,771$ are correct, and $31,306$ incorrect theorems are successfully repaired, yielding $100,077$ new theorems. For candidate tactic settings $4$ and $8$, the numbers of new theorems obtained are $22,691$ and $74,136$, respectively.  

After training models with the generated data, the second iteration produced $156,938$ new theorems for tactic settings $8$ and $160,389$ for $16$.  
The \textit{LeanComb}-Enhanced dataset, obtained by merging and deduplicating results from both iterations, contains $260,466$ new theorems.

%\vspace{-1em}

In the first iteration, the proportion of correct theorems after deduplication ranges from $17.5\%$ to $22.7\%$, increasing significantly to a maximum of $75.0\%$ in the second iteration. The proportion of newly generated theorems is also improved from $10.7\%, 20.3\%,$ and $16.9\%$  in the first iteration to $33.7\%, 41.7\%$, and $36.4\%$ in the second, demonstrating the effectiveness of iterative refinement. The $16$-tactic setting produces the highest number of theorems, while the $8$-tactic setting achieves the highest success rate for generating new theorems.

These results emphasize the importance of deduplication in reducing redundancy and improving data quality, while correction steps can improve the success rate. The observed improvement in the second iteration validates the effectiveness of iterative optimization in refining theorem generation.

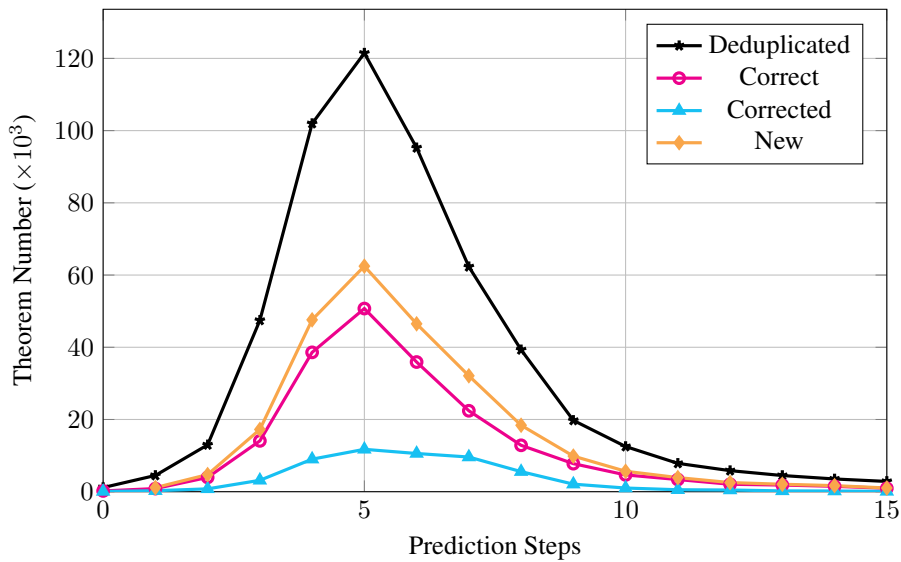
\begin{figure}[htp!]
\centering
\begin{tikzpicture}
\begin{axis}[
    width=12cm, height=8cm,
    xlabel={Prediction Steps},
    ylabel={Theorem Number ($\times 10^3$)},
    legend pos=north east,
    grid=major,
    xmin=0, xmax=15,  % 横坐标范围到15
    ymin=0,  % 纵坐标从0开始
    xtick={0,5,10,15},  % 横坐标刻度
    ytick distance=20, % 纵坐标步长
    scaled y ticks=false,
    ticklabel style={/pgf/number format/fixed}
]

% 第一条曲线: deduplicated number
\addplot[color=black, mark=star, very thick] coordinates {
    (0,1.143) (1,4.496) (2,12.974) (3,47.456) (4,101.950) (5,121.457) (6,95.244) (7,62.301)
    (8,39.295) (9,19.751) (10,12.512) (11,7.848) (12,5.831) (13,4.478) (14,3.525) (15,2.842)
};
\addlegendentry{Deduplicated}

% 第二条曲线: correct theorem number
\addplot[color=magenta, mark=o, very thick] coordinates {
    (0,0.209) (1,0.864) (2,3.951) (3,14.076) (4,38.588) (5,50.723) (6,35.911) (7,22.410)
    (8,12.846) (9,7.767) (10,4.646) (11,3.369) (12,2.060) (13,1.838) (14,1.499) (15,0.939)
};
\addlegendentry{Correct}

% 第三条曲线: corrected theorem number
\addplot[color=cyan!70, mark=triangle*, very thick] coordinates {
    (0,0.062) (1,0.273) (2,0.788) (3,3.156) (4,9.010) (5,11.736) (6,10.579) (7,9.600)
    (8,5.556) (9,2.089) (10,1.014) (11,0.541) (12,0.466) (13,0.224) (14,0.181) (15,0.084)
};
\addlegendentry{Corrected}

% 第四条曲线: new theorem number
\addplot[color=orange!80, mark=diamond*, very thick] coordinates {
      (1,1.137) (2,4.739) (3,17.232) (4,47.598) (5,62.459) (6,46.490) (7,32.100)
    (8,18.402) (9,9.856) (10,5.660) (11,3.910) (12,2.526) (13,2.062) (14,1.680) (15,1.023)
};
\addlegendentry{New} 
\end{axis}
\end{tikzpicture} 
%\vspace{-1em}
\caption{Distribution of Theorem Numbers on Prediction Steps}
\label{fig:1}
\end{figure}

\noindent
\textbf{Prediction Steps.}
Fig.\ref{fig:1} presents the distribution of theorem counts across different prediction steps, categorized into four groups: deduplicated, correct, corrected, and new theorems. The data reveal a peak around a prediction step of $6$, indicating that most theorems are concentrated at this length regardless of their classification. The number of deduplicated theorems exhibits the highest count, peaking at $121,457$. In contrast, the correct and new theorems follow similar trends but at lower magnitudes, suggesting that a significant proportion of generated theorems are either duplicates or require correction. 
The corrected theorems, represented by the green triangles, remain consistently lower than the other categories, highlighting the challenges in refining generated theorems through correction mechanisms.
 
%\vspace{-2.2em}

Additionally, the distribution demonstrates a sharp decline in theorem numbers beyond a prediction step of $6$, implying that longer proof sequences are less frequent and potentially more challenging to generate and verify. These findings suggest that optimizing theorem generation should focus on mid-range prediction steps, where the balance between uniqueness, correctness, and novelty is most favorable.
  % More details about \textit{LeanComb}-Enhanced dataset are provided in Appendix~\ref{AppendixD}.

\subsection{Theorem Proving on \textit{LeanComb} Benchmark and MiniF2F}
We evaluate the theorem-proving capabilities of LLMs on the test sets of  \textit{LeanComb} benchmark and miniF2F~\cite{zheng2022minif2f}, utilizing Pass@1 (\%) as the primary evaluation metric. Pass@1 is defined as the success rate of constructing a valid proof within a single attempt, constrained by a strict wall time limit of $10$ minutes. The experiment involves four models: Gemma2~\cite{gemma_2024}, Mathstral3~\cite{Mathstral}, Llama3~\cite{llama3modelcard}, and Mistral~\cite{jiang2023mistral}.  
The baseline models are initially pre-trained on Mathlib4 and subsequently fine-tuned on the \textit{LeanComb} training set. To further enhance their performance, an additional fine-tuning stage is conducted using an augmented dataset generated by our theorem generator. During inference, the number of candidate tactics generated by the models is restricted to a maximum of 16.

\textbf{Evaluation on \textit{LeanComb} Benchmark.}
We utilize the \textit{LeanComb} benchmark to optimize the tactic selection of LLMs, thereby substantially enhancing their theorem-proving capabilities. As presented in Table \ref{tab_3}, experimental results indicate that all models exhibit significant performance improvements after fine-tuning on the \textit{LeanComb} and \textit{LeanComb}-Enhanced datasets, with success rate increases ranging from $5\%$ to $17\%$. Notably, models trained on the first iteration of generated data (1st Iter) show substantial improvements, with success rates increasing by up to $13\%$ over the baseline. However, further refinement in the second iteration (2nd Iter) leads to even more significant gains, with a maximum success rate of $26\%$. This iterative process underscores the necessity of multiple fine-tuning stages to fully unlock the potential of the generated datasets.

These findings highlight the importance of continued training and the effectiveness of the generated datasets in advancing model performance in ATP tasks for combinatorial identities.

\begin{table}[htp!]
\centering
\caption{Success rates on the \textit{LeanComb} test set with Pass@1 ($\%$).}
\vspace{0.6em}
\label{tab_3}
\begin{tabular}{p{3cm}|>{\centering\arraybackslash}p{1.6cm} >{\centering\arraybackslash}p{1.8cm} >{\centering\arraybackslash}p{1.8cm} >{\centering\arraybackslash}p{1.8cm}}
\hline
\multirow{2}{*}{\centering\textbf{Model}} 
& \multirow{2}{*}{\centering\textbf{Base}} 
& \multirow{2}{*}{\centering\textbf{\textit{LeanComb}}} 
& \multicolumn{2}{c}{\textbf{\textit{LeanComb}-Enh.}} \\ 
\cmidrule {4-5} 
& & &  \textbf{1st Iter} & \textbf{2nd Iter} \\
\hline
Gemma2 - 9B & $10\%$ & $15\%$ & $17\%$  & $\textbf{23\%}$  \\ 
Mathstral3 - 8B &  $9\%$ & $19\%$ & $22\%$ & $\textbf{25\%}$  \\
Llama3 - 8B &  $9\%$ & $15\%$ & $17\%$ & $\textbf{22\%}$ \\ 
Mistral - 7B &  $9\%$ & $17\%$& $18\%$  & $\textbf{26\%}$   \\ 
\hline
\end{tabular}
\end{table}

\textbf{Evaluation on MiniF2F.}
Table~\ref{tab_4} provides a comparative analysis of different LLMs fine-tuned with \textit{LeanComb} and \textit{LeanComb}-Enhanced, on the miniF2F-test dataset. We evaluate the impact of our datasets on the success rates of fine-tuned models when tested on the miniF2F dataset. The success rate of the model trained on Mathlib+\textit{LeanComb}, is about $0.4\%$ to $4.1\%$ higher than that of the model trained solely on Mathlib, with an average improvement of $1.45\%$. Meanwhile, the success rate of the model trained on Mathlib+\textit{LeanComb}-Enhanced, shows an increase of about $3.3\%$ to $6.2\%$ compared to the model trained solely on Mathlib, with an average improvement of $4.9\%$.  

Notably, when we look at the performance of the models fine-tuned with \textit{LeanComb}-Enhanced in two iterations, significant improvements are observed across all models. The first iteration (1st Iter) provides an increase of around $1.0\%$ to $3.7\%$ over \textit{LeanComb}, and further fine-tuning in the second iteration (2nd Iter) leads to even more substantial improvements, with success rates rising by $1.1\%$ to $2.8\%$. For example, the Gemma2 model’s success rate increases from $27.9\%$ (1st Iter) to $30.3\%$ (2nd Iter), and similarly, the Mathtral3 model improves from $33.6\%$ to $36.1\%$. These results clearly demonstrate the effectiveness of the iterative fine-tuning process, with the second iteration playing a crucial role in achieving the highest performance across all models.

\begin{table}[htp!]
\centering
\caption{Success rates on the miniF2F test set with Pass@1($\%$).}
\vspace{0.6em}
\label{tab_4}
\begin{tabular}{p{3cm}|>{\centering\arraybackslash}p{1.6cm} >{\centering\arraybackslash}p{1.8cm} >{\centering\arraybackslash}p{1.8cm} >{\centering\arraybackslash}p{1.8cm}}
\toprule
\multirow{2}{*}{\centering\textbf{ Model}} 
& \multirow{2}{*}{\centering\textbf{Base}} 
& \multirow{2}{*}{\centering\textbf{\textit{LeanComb}}} 
& \multicolumn{2}{c}{\textbf{\textit{LeanComb}-Enh.}} \\ 
\cmidrule {4-5} 
& & &  \textbf{1st Iter} & \textbf{2nd Iter} \\

\midrule 
Gemma2 - 9B &  $26.2\%$ & $27.0\%$ & $27.9\%$ & $\textbf{30.3\%}$  \\ 
Mathtral3 - 8B & $29.9\%$ & $30.3\%$ & $33.6\%$ & $\textbf{36.1\%}$ \\ 
Llama3 - 8B &  $29.1\%$ &  $31.1\%$ & $34.0\%$ & $\textbf{34.8\%}$ \\ 
 Mistral - 7B &  $28.7\%$ & $32.8\%$ & $33.2\%$ & $\textbf{34.8\%}$  \\ 
\bottomrule
\end{tabular}
\end{table}

The above analysis can also support that our datasets are a necessary complement to enhance the automated theorem proving capabilities of existing LLMs. 

\section{Conclusion}
In this work, we presented a novel automated theorem generator, ATG4CI, for combinatorial identities, which started by calling LLM to yield the candidate tactics and then applied deep reinforcement learning for tactic prediction. We manually built \textit{LeanComb} benchmark for combinatorial identities and automatically generated the \textit{LeanComb}-Enhanced dataset obtained from ATG4CI. To evaluate its impact, we fine-tuned LLMs on these datasets. Experimental results indicate that our benchmark and dataset significantly improve the performance of LLMs in automated theorem proving.

 \newpage
\bibliographystyle{unsrt}  
\bibliography{LeanComb}

\nocite{langley00}

\newpage
\appendix
\onecolumn
  
%%%%%%%%%%%%%%%%%%%%%%%%%%%%%%%%%%%%%%%%%%%%%%%%%%%%%%%%%%%%%%%%%%%%%%%%%%%%%%%

\section{ Identities in \textit{LeanComb} Benchmark}
\label{AppendixA}

The \textit{LeanComb} benchmark dataset is constructed based on authoritative literature in the field of classical combinatorics~\cite{spivey2019art,gould1972combinatorial,shi2001combinatorial}, aiming to provide a high-quality mathematical reasoning benchmark to support automated theorem proving for combinatorial identities. The dataset comprises a total of 727 theorems, with the training set consisting of 627 theorems (including 418 combinatorial identities and 209 general theorems), while the test set contains 100 theorems. Additionally, we have formally introduced 9 new mathematical definitions, covering fundamental concepts such as Bell numbers, Stirling numbers of the first and second kinds, and combinations and permutations, thereby enhancing the expressiveness of the benchmark dataset. 

% Our \textit{LeanComb} test set and evaluation results are available at \href{https://anonymous.4open.science/r/LeanComb-2EB8/README.md}{Our LeanComb Repository}.

This dataset encompasses numerous classical theorems in combinatorics, including the Negative Binomial Series, Binet’s Formula, Vandermonde’s Identity, Trinomial Revision. For example:
\begin{align*}
 &\text{Negative Binomial Series:} \quad \forall x \in [0,1] \text{ and } n \in \mathbb{Z}_{\geq 0}, \quad \frac{1}{(1-x)^{n+1}} = \sum_{k=0}^{\infty} \binom{n+k}{n} x^k , \\
 &\text{Binet's Formula:} \quad F_n = \frac{\phi^n - \psi^n}{\sqrt{5}}.
\end{align*}

The expressions \( \binom{n}{k} \) and \( C_n^k \) are equivalent. To systematically illustrate the diversity of the dataset and the core mathematical structures involved, we have selected several representative combinatorial identity examples and categorized them based on their roles within the dataset.  
The selected examples are divided into two parts: one from the training set and the other from the test set. Each part includes various types of combinatorial identities, covering key topics such as the properties of Stirling numbers, classical combinatorial identities, and their variants. These examples not only provide an intuitive understanding of the dataset's composition but also establish a solid foundation for subsequent theoretical analysis and model evaluation.

\subsection{Training Set}

 \begin{itemize}
\setlength{\itemsep}{8pt}
\setlength{\parsep}{8pt}
\setlength{\parskip}{10pt}
 
    \item[] \textbf{idt\_15}
    
    Goal:
    \(
    \sum_{k=0}^{m}\binom{n+k}{k} = \binom{n+m+1}{m}
    \)

    \item[] \textbf{idt\_101}

    Goal:
    \(
    \sum_{k=0}^{n} \binom{n}{k}\frac{(x_{2}^{k+1}-x_{1}^{k+1})y^{n-k}}{(k+1)^2} = \frac{1}{n+1}\sum_{k=0}^{n}\frac{((x_2+y)^{k+1}-(x_1+y)^{k+1})y^{n-k}}{k+1}
    \)

    \item[] \textbf{idt\_105}

    Premises:
    \(a, b \in  \mathbb{R}_{\geq0}\)

    Goal:
    \(
    B(a,b) = \frac{\Gamma(a)\Gamma(b)}{\Gamma(a+b)}
    \)

    \item[] \textbf{idt\_106}

    Premises:
    \( n-k > -1 \)

    Goal:
    \(
    \frac{k!}{n^{\underline{k}}} = (n+1) \int_{0}^{1} x^k (1-x)^{n-k} dx
    \)

    \item[] \textbf{idt\_109}

    Premises:
    \( x > 0 \)

    Goal:
    \(
    \sum_{k=0}^{n} \binom{n}{k} \frac{(-1)^k}{k+x} = \frac{n!}{x(x+1)...(x+n)}
    \)

    \item[] \textbf{idt\_112}

    Goal:
    \(
    \sum_{k=0}^{n} \binom{n}{k} k (k+m) = n (n + 2m + 1) 2^{n-2}
    \)

    \item[] \textbf{idt\_115}

    Goal:
    \(
    \sum_{k=0}^{r} \binom{n}{r-k} \binom{m+k}{m} (-1)^k = \binom{n-m-1}{r}
    \)

    \item[] \textbf{idt\_117}

    Goal:\(
   \sum_{k=0}^{n} \binom{n}{k}\frac{ (n + 1) (n + 2) (n + 3)}{(k+1)(k+2)(k+3)}  = 2^{n+3} - 1 - (n+3) - \frac{(n+2)(n+3)}{2}\)

    \item[] \textbf{idt\_121}

    Goal:
    \(
    \sum_{n=0}^{\infty} \frac{2^n}{(2n+1) \binom{2n}{n}} = \frac{\pi}{2}
    \)

    \item[] \textbf{idt\_124}

    Goal:
    \(
    \frac{1}{\binom{n}{k}} = 2 (n+1) \int_{0}^{\frac{\pi}{2}} \sin^{2k+1} \theta \cos^{2n-2k+1} \theta d\theta
    \)

    \item[] \textbf{idt\_131}

    Premises:
    \( 0 \leq p \leq 1 \)

    Goal:
    \(
    \sum_{k=0}^{n} \binom{n}{k}  p^k (1-p)^{n-k} k^2 = n^2 p^2 + n p (1-p)
    \)
    \item[] \textbf{idt\_132}

    Goal:
    \(
    \sum_{k=0}^{m} \binom{m}{k} \binom{n}{r-k} k = \binom{m+n}{r}\frac{rm}{m+n}
    \)
    
    \item[] \textbf{idt\_179}

    Premises:
    \( n \leq m \)

    Goal:
    \(
    \Delta^{n}F_{m} = F_{m-n}
    \)

    \item[] \textbf{idt\_212}

    Goal:
    \(
    \sum_{k=1}^{n} \binom{n}{k}\frac{(-1)^k}{k(k+1)\cdots(k+m)}=-\frac{H_{n+m} - H_{m}}{m!}
    \)
    
    \item[] \textbf{idt\_285}

    Premises:
    \( 0 < y \)

    Goal:
    \(
    \sum_{k=0}^{n}\binom{n}{k}i^{k}y^{n-k} = (y^2+1)^{n/2}e^{i \mkern3mu n \arctan(1/y)}
    \)

    \item[] \textbf{idt\_318}
    
    Goal:
    \(
    \sum_{k\geq 0}\binom{n}{2k+1}3^k(-1)^k = \frac{2^n}{\sqrt{3}} \sin \frac{n\pi}{3} 
    \)

\end{itemize}

\subsection{Test Set}

 \begin{itemize}
\setlength{\itemsep}{8pt}
\setlength{\parsep}{8pt}

    \item[]   \textbf{test\_002}

Goal: \( n^{\overline{k+1}} = (k + 1)! \;\times\; \binom{n + k}{k + 1} \)

\item[] \textbf{test\_005}

Premises: \( n : \mathbb{N}; n \ge 1 \)

Goal: \( \sum_{k=1}^{n}
   \frac{1}{k}\,\binom{2(k-1)}{k-1}\,\binom{2(n-k)}{n-k}
   \;=\;\frac{1}{2}\,\binom{2n}{n} \)

\item[] \textbf{test\_011}

Premises: \( n,m:\mathbb{N};m \ge 2n;1 \le n \)

Goal: \( \displaystyle \sum_{k=0}^{n} (-1)^{k}\,\binom{n}{k}\,\binom{m - 2k}{\,n - 1} \;=\; 0.  \)

\item[] \textbf{test\_025}

Goal: \( \sum_{k=0}^{\lfloor n/2 \rfloor} \binom{n-k}{k} 
   \;=\;\frac{1}{\sqrt{5}}
        \Bigl(\Bigl(\frac{1 + \sqrt{5}}{2}\Bigr)^{\,n + 1}
             \;-\;\Bigl(\frac{1 - \sqrt{5}}{2}\Bigr)^{\,n + 1}\Bigr).  \)

\item[] \textbf{test\_037}

Goal: \( \sum_{k=0}^{n} (-1)^{\,k} \,\binom{n}{k}\,\binom{m - k}{r}
   \;=\;\binom{m - n}{\,r - n}. \)

\item[] \textbf{test\_054}

Premises: \( 1 \le n \)

Goal: \( \sum_{k=0}^{m} \binom{k + n - 1}{n - 1}
\;=\;\binom{n + m}{n}. \)

\item[] \textbf{test\_069}

Premises: \( n \ge 3 \)

Goal: \( n^3 \;=\; 6\,\binom{n}{3} \;+\; 6\,\binom{n}{2} \;+\; \binom{n}{1} \)

\item[] \textbf{test\_076}

Goal: \( \sum_{k=0}^{n-1} (-1)^k \,\Bigl(\cos\!\bigl(\tfrac{k\pi}{n}\bigr)\Bigr)^n
   \;=\;\frac{n}{2^{\,n-1}}\,.  \)

\item[] \textbf{test\_088}

Premises: \( 1 \leq n \)

Goal: \( \sum_{k=0}^{n} \binom{2n}{k} = 2^{2n - 1} + \frac{1}{2} \binom{2n}{n} \)

\item[] \textbf{test\_100}

Premises: \( k \leq n \)

Goal: \( \binom{n}{k} \cdot \mathrm{B}(k+1, n-k+1) = \frac{1}{n+1} \)

\end{itemize}

\section{Lean4Kit}
\label{AppendixB}

In our automated theorem generator ATG4CI, Lean4Kit plays a key role in data extraction and interaction. During the experimental evaluation and testing phase, Lean4Kit assists LLMs in theorem proving through interaction with Lean 4. In fact, given any repo in Lean 4, our offline toolkit can convert Lean files into JSON format data, extracting all state-tactic pairs.

\subsection{Data Extraction from Lean Codes}

The complete Lean code (including imports) is converted into a JSON-formatted data structure tree (infotree) in the static extraction process. The conversion function   \textit{run\_all\_tactics(self, code, env=None, verbose=True)}  returns data in a format that includes the tactic, as well as the before-and-after states of the goals (goalsBefore and goalsAfter), for example:

\begin{tcolorbox}[
  colback=SkyBlue!5, %  背景
  colframe= black, %  框线
  % title=\textit{ },
  % fonttitle=\bfseries\footnotesize,
  % coltitle=titlecolor,
  rounded corners,
  boxrule=0.6mm
]
\noindent\textbf{pp}: rw [abelidentity\_eq\_add] \\
\textbf{ name }:  Lean.Parser.Tactic.rwSeq \\
\textbf{ goalsBefore :}
\begin{lstlisting}
n : ℕ
x y : ℝ
hn1 : 1 ≤ n
hx : x ≠ 0
hy : y ≠ 0 
⊢ abelidentity x y (-1) (-1) n = (1 / x + 1 / y) * (x + y + ↑n) ^ (n - 1)
\end{lstlisting}
\textbf{ goalsAfter :}
\begin{lstlisting} 
n : ℕ
x y : ℝ
hn1 : 1 ≤ n
hx : x ≠ 0
hy : y ≠ 0
⊢ abelidentity x (y + 1) (-1) (-1 + 1) (n - 1) + abelidentity (x + 1) y (-1 + 1) (-1) (n - 1) = (1 / x + 1 / y) * (x + y + ↑n) ^ (n - 1)

case hn
n : ℕ
x y : ℝ
hn1 : 1 ≤ n
hx : x ≠ 0
hy : y ≠ 0
⊢ n ≥ 1
\end{lstlisting}
\end{tcolorbox}

\subsection{Dynamic Interaction with Lean for Theorem Proving}

In the dynamic interaction process, we interact with Lean to perform automated theorem proving. The process mainly relies on the following functions:
\begin{itemize}
    \item   \textit{run\_import}(self, code, env=None, verbose=False) : Used to import necessary environments and dependencies.
    \item   \textit{new\_thm}(self, code, env=None, verbose=False) : Generates a new theorem based on the provided theorem description.
\end{itemize}
 
The provided code parameter represents the description of the initial theorem, for example:

 \vspace{.2em}
\begin{lstlisting}
theorem idt_84(n: ℕ)(h: m < n): ∑ k in range (n + 1), n.choose k (n - k)^m (-1 : ℝ)^k = 0 := by sorry
\end{lstlisting}
\vspace{.2em}

This function returns an initial state for subsequent tactic applications:
\begin{itemize}
  \item    \textit{run\_tactic}(self, tactic, proofState, cmd\_type= 'tactic', verbose=False): Executes a tactic and returns the new state after the tactic is applied.
 \item   \textit{run\_have\_tactic}(self, tactic, proofState, cmd\_type='have', verbose=False) : Executes a “have” tactic.
\end{itemize}

During the interaction, we can also use function $is\_correct\_and\_finished(self, code, verbose=False, timeout=160)$  to check if the theorem is correct and whether the proof is complete, with the judgment based on the information view (Lean Infoview) on the right side.

\begin{tcolorbox}[
  colback=SkyBlue!5, %  背景
  colframe= black, %  框线
  % title=\textit{ },
  % fonttitle=\bfseries\footnotesize,
  % coltitle=titlecolor,
  rounded corners,
  boxrule=0.6mm
]
\textbf{ ProcessingCmd :} 
\begin{lstlisting}
example (a b c : 2115)(h : a = b): a ^ 2 + c= b ^ 2 + c:= by sorry 
“env”: 0
\end{lstlisting} 
\textbf{proofstates :} [0],\\
\textbf{goals :}  
\begin{lstlisting}
 a b c : N  h : a = b ⊢ a ^ 2 + c = b ^ 2 + c   
\end{lstlisting} 
\textbf{error}: False\\
\textbf{messages}: [\{'severity': 'warning', 'pos': \{'line': 1,'column': 0\}, 'endPos': \{'line': 1,'column': 7\}, 'data': “declaration uses 'sorry'”\}] \\
\textbf{sorries}: [\{'proofState' : 0, 'pos': \{'line' : 1,'column' : 58\}, 'goals', 'endPos' : \{'line': 1,'column': 63\}\}]\\
\textbf{is finish}: False
   
\end{tcolorbox}
\vspace{.2em}

The returned  Tactic State  format includes the following key fields:
\begin{itemize}
 \item \textit{messages}: Contains information during the interaction, such as “no goals,” “declare use sorry,” “unknown tactic,” etc.
 \item  \textit{proofstates} : An integer list uniquely identifies the current state, where each integer corresponds to a subgoal’s state.
 \item \textit{goals} : The current subgoals.
 \item \textit{error}, \textit{finishFlag} : These parameters are assigned by analyzing messages . The error indicates whether an error occurred (True for error), and finishFlag indicates whether the proof is complete (True for completed proof).
 \end{itemize}

\section{An Illustrative Example for ATG4CI}
 \label{AppendixC}

For the \textit{LeanComb} benchmark, data augmentation is performed using our theorem generator, ATG4CI. Some theorems in the dataset are augmented to generate up to 4,000 new instances in one iteration, resulting in a total of 260,466 newly generated theorems. Among these, 56,747 erroneous theorems are successfully corrected. Below, we demonstrate the workflow of ATG4CI using a typical example, which corresponds to the mathematical formula $\sum_{k=0}^{n}\binom{n}{k}F_{m+k} = F_{2n+m}$:

\vspace{0.6em}
\begin{lstlisting}
import Mathlib
import Theorem.valid.idt_179

open Finset Nat

theorem idt_182 (m n : ℕ) : ∑ k in Finset.range (n + 1), Nat.choose n k * Nat.fib (m + k) = Nat.fib (2 * n + m) := by
  suffices ∑ k in Finset.range (n + 1), Nat.choose n k * Nat.fib (m + k) = (Nat.fib (2 * n + m) : ℝ) by
    norm_cast at this
  let g := fun k => Nat.fib (m + n + k)
  -- $\sum_{k=0}^{n} \binom{n}{k} * fib(m+k) = \sum_{k=0}^{n} \binom{n}{k} * fib(m+n-n+k)$
  have h₁: ∑ k in Finset.range (n + 1), Nat.choose n k * Nat.fib (m + k) =
    ∑ k in Finset.range (n + 1), Nat.choose n k * Nat.fib (m + n - n + k) := by
    congr! 1 with k hk
    simp at hk
    congr; omega
  rw [h₁]
  -- $\sum_{k=0}^{n} \binom{n}{k} * fib(m+n-n+k) = \sum_{k=0}^{n} \binom{n}{k} * fib(m+n-k)$
  rw [← Finset.sum_flip]
  have h₂: ∑ k ∈ range (n + 1), n.choose (n - k) * fib (m + n - n + (n - k)) = ∑ k ∈ range (n + 1), n.choose k * fib (m + n - k) := by
    congr! 1 with k hk
    simp at hk
    rw [choose_symm (by linarith)]
    congr; omega
  rw [h₂]
  -- $\sum_{k=0}^{n} \binom{n}{k} * \text{fib}(m+n-k) = \sum_{k=0}^{n} \binom{n}{k} * \sum_{j=0}^{k} \binom{k}{j} * g(j)*(-1)^k*(-1)^j$
  
  ... ...
  
  have h₇: ∑ k ∈ range (n + 1), n.choose k * g k * (-1) ^ k * ∑ x ∈ range (n - k + 1), (n - k).choose (x) * (-1 : ℝ) ^ (x + k) =
    ∑ k ∈ range (n + 1), n.choose k * g k * ∑ x ∈ range (n - k + 1), (n - k).choose (x) * (-1 : ℝ) ^ x := by
    congr! 1 with k _
    rw [mul_assoc, mul_sum]
    congr 1
    congr! 1 with j _
    rw [mul_comm, pow_add, ← mul_assoc, mul_assoc, ← pow_add, ← two_mul, pow_mul]
    ring
  rw [h₇, ← Finset.sum_range_add_sum_Ico _ (m := n) (by omega), show 2 * n + m = m + n + n by omega]
  simp [g]
  apply sum_eq_zero
  intro k hk
  rw [_root_.mul_eq_zero]
  right
  simp at hk
  rw [show (0 : ℝ) = (-1 + 1) ^ (n - k) by simp; rw [zero_pow (by omega)], add_pow]
  simp [mul_comm]
\end{lstlisting}

Combinatorial Identity $182$ encompasses key concepts such as binomial coefficients, Fibonacci numbers, and generating functions. The proof consists of $77$ lines involving $23$ distinct tactics, resulting in a proof length and tactic diversity of $23$. Structurally, it forms a $24$-layer proof tree, from which $22$ distinct \textit{P3s} can be extracted. The longest and shortest \textit{P3s} are highlighted as follows:

\begin{lstlisting}
theorem P3s_1 (m n : ℕ) : ∑ k in Finset.range (n + 1), Nat.choose n k * Nat.fib (m + k) = Nat.fib (2 * n + m) := by
  suffices ∑ k in Finset.range (n + 1), Nat.choose n k * Nat.fib (m + k) = (Nat.fib (2 * n + m) : ℝ) by
    norm_cast at this
  ... ...
  rw [h₁]
\end{lstlisting}

\begin{lstlisting}
theorem P3s_22 (m n : ℕ) : ∑ k in Finset.range (n + 1), Nat.choose n k * Nat.fib (m + k) = Nat.fib (2 * n + m) := by
  suffices ∑ k in Finset.range (n + 1), Nat.choose n k * Nat.fib (m + k) = (Nat.fib (2 * n + m) : ℝ) by
    norm_cast at this
  ... ...
  rw [show (0 : ℝ) = (-1 + 1) ^ (n - k) by simp; rw [zero_pow (by omega)], add_pow]
\end{lstlisting}

Based on these \textit{P3s}, our RLSCI framework facilitates tactic prediction, enabling the generation of candidate theorems. We further construct their proofs, followed by deduplication and correctness verification. Below, we present a selection of intriguing combinatorial identities generated by our theorem generator using the aforementioned examples:

\begin{itemize}
\setlength{\itemsep}{8pt}
\setlength{\parsep}{8pt}
\setlength{\parskip}{10pt}
 
    \item[] \textbf{idt\_182\_3\_12:}
  $$\sum_{k=0}^{n}(\binom{n}{k}\sum_{j=0}^{k}\binom{k}{j}F_{m + n + j}(-1)^{j})(-1)^{k}=F_{2n+m}
 $$
  \item[] \textbf{idt\_182\_1\_3:} 
  $$\sum_{k=0}^{n}\binom{n}{k}F_{m+n-k}=F_{2n+m}$$
  \item[] \textbf{idt\_182\_0\_5:} $$\sum_{x=0}^{n}\sum_{k=0}^{x}\binom{n}{k}\binom{n-k}{x-k}F_{m + n + k}(-1)^{x+k}=F_{2n+m}$$
  \item[] \textbf{idt\_182\_0\_72} 
  $$\sum_{k=0}^{n}\binom{n}{k}F_{m+n-k}=F_{2n+m}$$

 \item[]\textbf{idt\_182\_6\_5}
$$ 
\sum_{k=0}^{n}\binom{n}{k}F_{m + n + k}(-1) ^ {k}\sum_{x=0}^{n-k}\binom{n-k}{x}(-1) ^ {x+k}= F_{2n + m}
  $$ 

 \item[] \textbf{idt\_182\_6\_23}
  $$\sum_{k=0}^{n-1}\binom{n}{k}F_{m + n + k}\sum_{j=0}^{n-k}\binom{n-k}{j}(-1)^j=0$$
  
 \item[]\textbf{idt\_182\_6\_49} 
     $$
     \binom{n}{k}F_{m + n + k}\sum_{x=0}^{n-k}\binom{n-k}{x}(-1)^x=0$$
     
  \item[]\textbf{idt\_182\_2\_52}
     $$\sum_{x=0}^{n-k}\binom{n-k}{x}(-1)^{x}=0
     $$ 
    \item[] \textbf{idt\_182\_2\_45}
    $$\sum_{r=0}^{n}\binom{n}{n-r}F_{m+(n-r)} = F_{2n+m}$$\vspace{0.3em}

\end{itemize}

\section{Dataset Statistic}
\label{AppendixD}

\subsection{Proof Steps}

We analyzed the proof steps of the \textit{LeanComb} benchmark and the enhanced dataset on a per-tactic basis and visualized their distributions in Fig. 3 and 4. Proof steps per tactic serve as an indicator of theorem complexity. Notably, as proof steps increase, the distribution of theorem steps becomes less concentrated. This phenomenon may result from the broader range of theorem steps observed in more complex cases. For instance, short proof steps (e.g., between $1$ and $10$) correspond to a higher number of theorems, which are more densely distributed. In contrast, when the proof steps exceed $40$, the number of theorems decreases significantly, and their distribution becomes sparser.

Specifically, in the \textit{LeanComb} benchmark, the maximum proof steps reach $166$, with $10$ theorems having proof steps exceeding $100$ lines. This suggests that the \textit{LeanComb} benchmark contains a small number of highly complex theorems. In contrast, the maximum number of proof steps in the enhanced dataset is $124$, with  $800$ theorems having proof steps exceeding $100$ lines. This indicates that the enhanced dataset contains a more significant number of long-proof theorems, reflecting a broader range of difficulty levels. As a result, the enhanced dataset offers greater diversity in terms of theorem complexity, potentially presenting a more comprehensive challenge for reasoning systems.

Furthermore, these differences suggest that while the \textit{LeanComb} benchmark includes a few highly complex theorems, its overall complexity distribution remains relatively concentrated. On the other hand, the enhanced dataset provides more extensive coverage of longer proofs, which may offer a more thorough evaluation of reasoning tactics regarding adaptability and generalization capabilities.

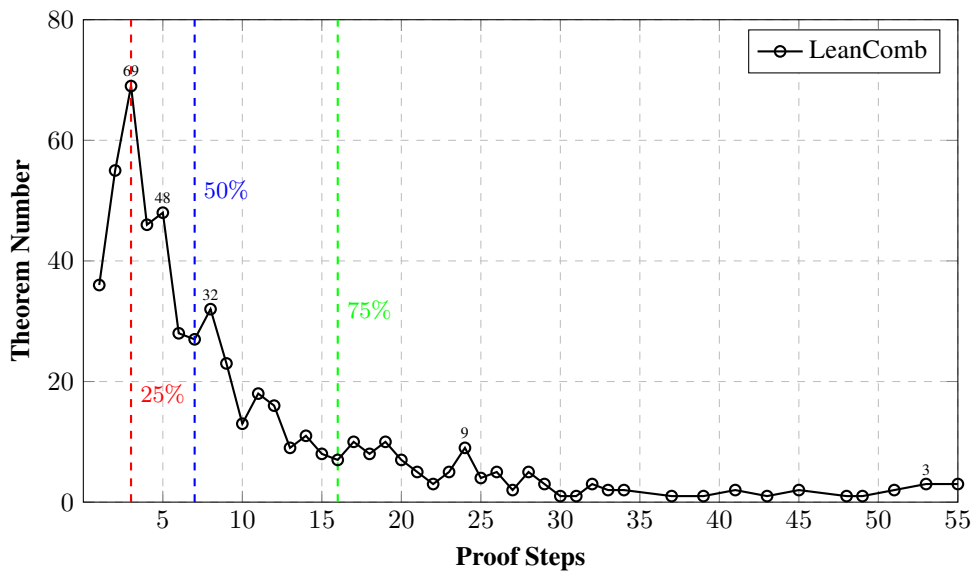
\begin{figure}[htp!]
    \centering
    \begin{tikzpicture}
        \begin{axis}[
            title={ },
            title style={font=\bfseries},
            xlabel={\textbf{Proof Steps}},
            ylabel={\textbf{Theorem Number}},
            xtick={5,10,15,20,25,30,35,40,45,50,55,60},
            ymin=0,
            ymax=80,
            xmin=0,
            xmax=55,
            grid=both,
            major grid style={dashed,gray!50},
            minor grid style={dotted,gray!20},
            width=0.8\textwidth,
            height=8cm,
        ]
        \addplot[color=black, mark=o,  thick] coordinates {(1,36)
              (2,55) (3,69) (4,46) (5,48) (6,28) (7,27) (8,32) (9,23) (10,13) 
            (11,18) (12,16) (13,9) (14,11) (15,8) (16,7) (17,10) (18,8) (19,10) (20,7) 
            (21,5) (22,3) (23,5) (24,9) (25,4) (26,5) (27,2) (28,5) (29,3) (30,1) 
            (31,1) (32,3) (33,2) (34,2) (37,1) (39,1) (41,2) (43,1) (45,2) (48,1) 
            (49,1) (51,2) (53,3)  (55,3)
        };
        \legend{LeanComb}
         % 手动添加标注
    \node[above, font=\tiny] at (axis cs: 3,69 ) {69};
    \node[above, font=\tiny] at (axis cs:5,48) {48};
    \node[above, font=\tiny] at (axis cs:8,32) {32};
    \node[above, font=\tiny] at (axis cs:24,9) {9};
      
    \node[above, font=\tiny] at (axis cs:(53,3) {3};
     \addplot[dashed, thick, red] coordinates {(3,0) (3,80)}; % 25%
        \node[anchor=west, red, font=\small] at (axis cs:3, 18) {$25\%$};

        \addplot[dashed, thick, blue] coordinates {(7,0) (7,80)}; % 50%
        \node[anchor=north west, blue, font=\small] at (axis cs:7, 55) {$50\%$};

        \addplot[dashed, thick, green] coordinates {(16,0) (16,80)}; % 75%
        \node[anchor=north west, green, font=\small] at (axis cs:16, 35) {$75\%$};
        \end{axis}
    \end{tikzpicture}
   
    \caption{Theorem count distribution across proof steps in \textit{LeanComb} benchmark.}
     \label{fig6}
\end{figure}

\begin{figure}[htp!]
    \centering
    \begin{tikzpicture}
        \begin{axis}[
            title={ },
            title style={font=\bfseries},
            xlabel={\textbf{Proof Steps}},
            ylabel={\textbf{Theorem Number ($\times 10^{3}$)}},
            xtick={5,10,15,20,25,30,35,40},
            ymin=0, 
            ymax=3,
            xmin=0,
            enlarge x limits=0.02,
            grid=both,
            major grid style={dashed,gray!50},
            minor grid style={dotted,gray!20},
            width=0.8\textwidth,
            height=8cm,
            font=\small,
        ]
        % 数据绘制
        \addplot[color=black, mark=square, thick] coordinates {
            (1,0.391) (2,0.171) (3,0.627) (4,1.679) (5,2.564) (6,2.374) (7,2.028) (8,1.613) (9,1.267) (10,1.129)
            (11,0.910) (12,0.750) (13,0.614) (14,0.519) (15,0.414) (16,0.488) (17,0.317) (18,0.289) (19,0.173) (20,0.097)
            (21,0.067) (22,0.022) (23,0.042) (24,0.031) (25,0.031) (26,0.034) (27,0.029) (28,0.047) (29,0.021) (30,0.012)
            (31,0.010) (32,0.009) (33,0.029) (35,0.008) (36,0.006) (37,0.008) (38,0.008) (40,0.001) 
        };
        \legend{\textit{LeanComb}-Enhanced}
        
        % 添加垂直线
        \addplot[dashed, thick, red] coordinates {(5,0) (5,6)}; % 25%
        \node[anchor=west, red, font=\small] at (axis cs:5, 1.6) {$25\%$};

        \addplot[dashed, thick, blue] coordinates {(8,0) (8,6)}; % 50%
        \node[anchor=north west, blue, font=\small] at (axis cs:8, 2.7) {$50\%$};

        \addplot[dashed, thick, green] coordinates {(16,0) (16,6)}; % 75%
        \node[anchor=north west, green, font=\small] at (axis cs:16, 2.7) {$75\%$};

        % 标注重要的点
        \node[above, font=\tiny] at (axis cs: 1,0.391 ) {0.391};
        \node[above, font=\tiny] at (axis cs: 5,2.564) {2.564};
        \node[above, font=\tiny] at (axis cs: 16,0.488) {0.488};
        \node[above, font=\tiny] at (axis cs: 28,0.047) {0.047};
        \node[above, font=\tiny] at (axis cs: 35,0.008) {0.008};
        \end{axis}
    \end{tikzpicture}
    \caption{Theorem count distribution across proof steps in \textit{LeanComb}-Enhanced dataset.}
    \label{fig7}
\end{figure}
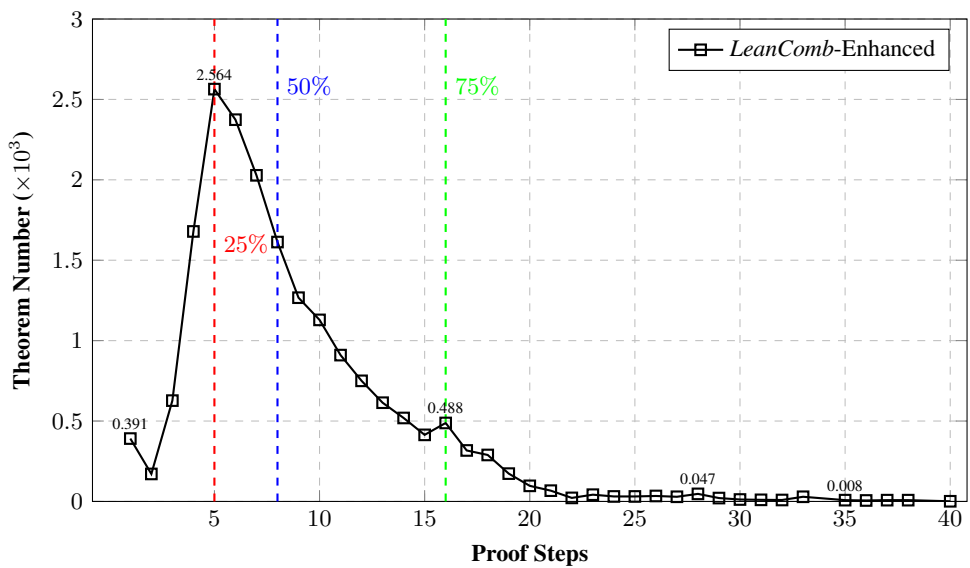
\vspace{.5em}

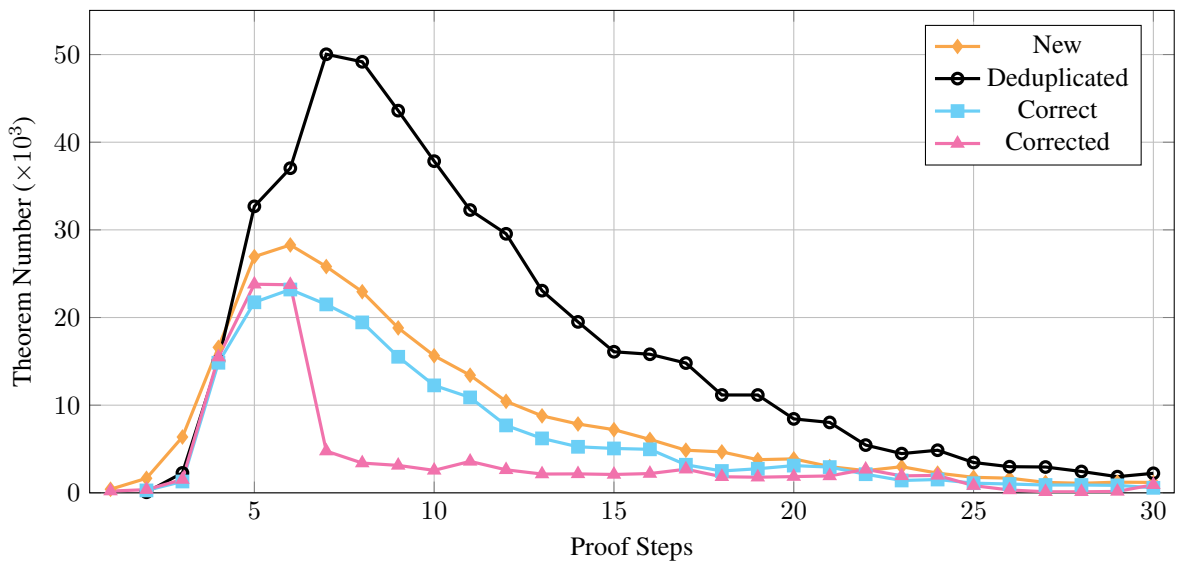
\begin{figure}[htp!]
\centering
\begin{tikzpicture}
\begin{axis}[  
    width=16cm, height=8cm,
    xlabel={Proof Steps},
    ylabel={Theorem Number ($\times 10^3$)},
    legend pos=north east,
    grid=major,
    xmax=30,  % 横坐标范围到30
    ymin=0,  % 纵坐标从0开始
    enlarge x limits=0.02,
    xtick={ 5,10,15,20,25,30},  % 横坐标刻度
    ytick distance=10, % 纵坐标步长
    scaled y ticks=false,
    ticklabel style={/pgf/number format/fixed}  ]

% 第一条曲线: 新定理数据
\addplot[color=orange!80, mark=diamond*,very thick] coordinates {
    (1,0.408) (2,1.657) (3,6.370) (4,16.597) (5,26.944) (6,28.278) (7,25.815) (8,22.942) 
    (9,18.814) (10,15.640) (11,13.417) (12,10.439) (13,8.781) (14,7.852) (15,7.205) 
    (16,6.097) (17,4.863) (18,4.671) (19,3.781) (20,3.876) (21,2.992) (22,2.528) 
    (23,2.981) (24,2.245) (25,1.770) (26,1.672) (27,1.189) (28,1.089) (29,1.217) (30,1.188)
};
\addlegendentry{New}

% 第一条曲线: deduplicated theorem number (调整为除以1000)
\addplot[color=black, mark=o,very thick] coordinates {
    (2,0.065) (3,2.272) (4,15.049) (5,32.683) (6,37.035) (7,50.034) (8,49.174) (9,43.602) 
    (10,37.840) (11,32.266) (12,29.551) (13,23.065) (14,19.501) (15,16.094) (16,15.809) 
    (17,14.804) (18,11.165) (19,11.162) (20,8.435) (21,8.032) (22,5.448) (23,4.476) 
    (24,4.860) (25,3.461) (26,2.984) (27,2.946) (28,2.437) (29,1.845) (30,2.227)
};
\addlegendentry{Deduplicated}

% 第二条曲线: correct theorem number (调整为除以1000)
\addplot[color=cyan!50, mark=square*,very thick] coordinates { 
    (2,0.275) (3,1.295) (4,14.839) (5,21.739) (6,23.196) (7,21.495) (8,19.445) (9,15.522) 
    (10,12.258) (11,10.884) (12,7.691) (13,6.209) (14,5.252) (15,5.063) (16,4.965) 
    (17,3.203) (18,2.492) (19,2.748) (20,3.107) (21,2.937) (22,2.125) (23,1.409) 
    (24,1.520) (25,1.103) (26,1.007) (27,0.892) (28,0.892) (29,0.853) (30,0.572)
};
\addlegendentry{Correct}

% 第三条曲线: corrected theorem number (调整为除以1000)
\addplot[color=magenta!70, mark=triangle*,very thick] coordinates { 
    (1,0.2) (2,0.362) (3,1.517) (4,15.485) (5,23.798) (6,23.739) (7,4.763) (8,3.397) 
    (9,3.129) (10,2.565) (11,3.588) (12,2.625) (13,2.143) (14,2.161) (15,2.095) 
    (16,2.199) (17,2.695) (18,1.835) (19,1.783) (20,1.853) (21,1.927) (22,2.682) 
    (23,1.937) (24,1.987) (25,0.822) (26,0.327) (27,0.121) (28,0.120) (29,0.160) (30,0.923)
};
\addlegendentry{Corrected}

\end{axis}
\end{tikzpicture}

\caption{Theorem distribution by proof Steps}
\label{fig:theorem_distribution2}
\end{figure}

\textbf{Distribution of Theorems in \( E^* \) by Theorem Type Count.}
Fig. \ref{fig:theorem_distribution2} shows the distribution of deduplicated, correct, corrected, and new theorems by proof steps in the enhanced dataset \(E^*\), including newly added data. Deduplicated theorems peak at proof step 6 with approximately $50,000$ and decline as steps increase, indicating shorter proofs dominate. Correct theorems peak slightly earlier at step 5 with $22,000$, showing that many deduplicated theorems fail correctness checks. Corrected theorems are fewer and decrease rapidly after step 5, suggesting that errors in longer proofs are harder to address. The new data aligns with these trends, peaking at step 6 with $28,000$ theorems but maintaining a steady count for longer proofs, highlighting improved generation of extended proofs yet persistent challenges in verification.  

These results emphasize two key challenges in automated theorem proving: the gap between deduplicated and correct theorems, particularly for longer proofs, underscores the need for more robust verification methods, and the rapid decline in corrected theorems highlights the difficulty of resolving errors in complex proofs. Improving proof generation and error correction tactics—especially for longer proofs—remains a crucial direction, alongside integrating advanced validation mechanisms to enhance correctness and diversity.

\subsection{Tactic Types Distribution}

This section presents a statistical analysis of the distribution of two datasets based on the number of tactic types involved in the proof of each theorem. A greater number of tactic types employed in the proof of a theorem generally indicates a higher level of complexity in the proof, as well as reflecting the diversity and depth of the problem. Moreover, the more tactic types used in the generated theorems, the stronger the model’s ability to explore and apply different tactics.

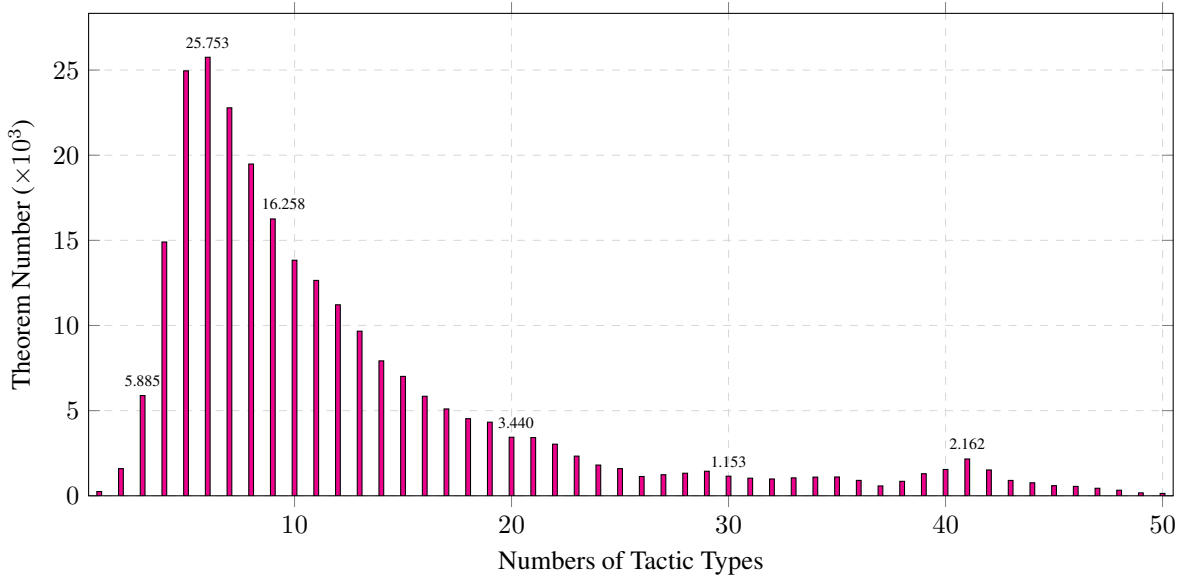
\begin{figure}[htp!]
\centering
\begin{tikzpicture}
    \begin{axis}[
        ybar,
        xtick={0,10,...,50}, % 设置横坐标刻度为0, 10, 20,...直到50
        xmax=50, % 将横坐标最大值限制为50
        ymin=0, % 设置纵坐标最小值为0
        enlarge x limits=0.01, % 调整柱子间距
        ylabel={Theorem Number ($\times 10^3$)},
        xlabel={Numbers of Tactic Types},
        grid=major,
        major grid style={dashed, gray!30},
        bar width=1.8pt,
        width=16cm, height=8cm
    ]
    \addplot[fill=magenta] coordinates {
        (1,0.249)(2,1.596)(3,5.885)(4,14.905)(5,24.954)(6,25.753)(7,22.782)(8,19.482)(9,16.258)(10,13.834)
        (11,12.648)(12,11.216)(13,9.668)(14,7.923)(15,7.012)(16,5.845)(17,5.101)(18,4.529)(19,4.323)(20,3.440)
        (21,3.420)(22,3.031)(23,2.330)(24,1.805)(25,1.597)(26,1.133)(27,1.237)(28,1.326)(29,1.442)(30,1.153)
        (31,1.034)(32,0.989)(33,1.053)(34,1.097)(35,1.106)(36,0.905)(37,0.578)(38,0.850)(39,1.296)(40,1.546)
        (41,2.162)(42,1.517)(43,0.903)(44,0.764)(45,0.595)(46,0.553)(47,0.443)(48,0.328)(49,0.172)(50,0.140)
    };
    
    % 手动添加标注
    
    \node[above, font=\tiny] at (axis cs:3,5.885) {5.885};
    \node[above, font=\tiny] at (axis cs:6,25.753) {25.753};
    \node[above, font=\tiny] at (axis cs:9.5,16.258) {16.258};
    \node[above, font=\tiny] at (axis cs:20.2,3.440) {3.440};
    \node[above, font=\tiny] at (axis cs:30,1.153) {1.153};
    \node[above, font=\tiny] at (axis cs:41,2.162) {2.162};

    \end{axis}
\end{tikzpicture}
\caption{Theorem Distribution by Tactic Types of Enhanced Dataset $E^*$}
\label{fig:theorem_distribution3}
\end{figure}

Figure \ref{fig:theorem_distribution3} illustrates the distribution of theorem counts as a function of the number of tactic types in the enhanced dataset $E^*$. The theorem count peaks at 25,753 when the number of tactic types reaches six, after which a gradual decline is observed as the number of tactic types increases. This decline becomes particularly steep when the number of tactic types exceeds 20. This trend suggests that most theorems can be proven using a relatively small set of tactic combinations, reflecting the dominance of simpler theorems in combinatorial mathematics. However, the sparsity of theorems requiring many tactic types points to an underrepresentation of complex theorems in the dataset, presenting a significant challenge for automated theorem-proving systems.

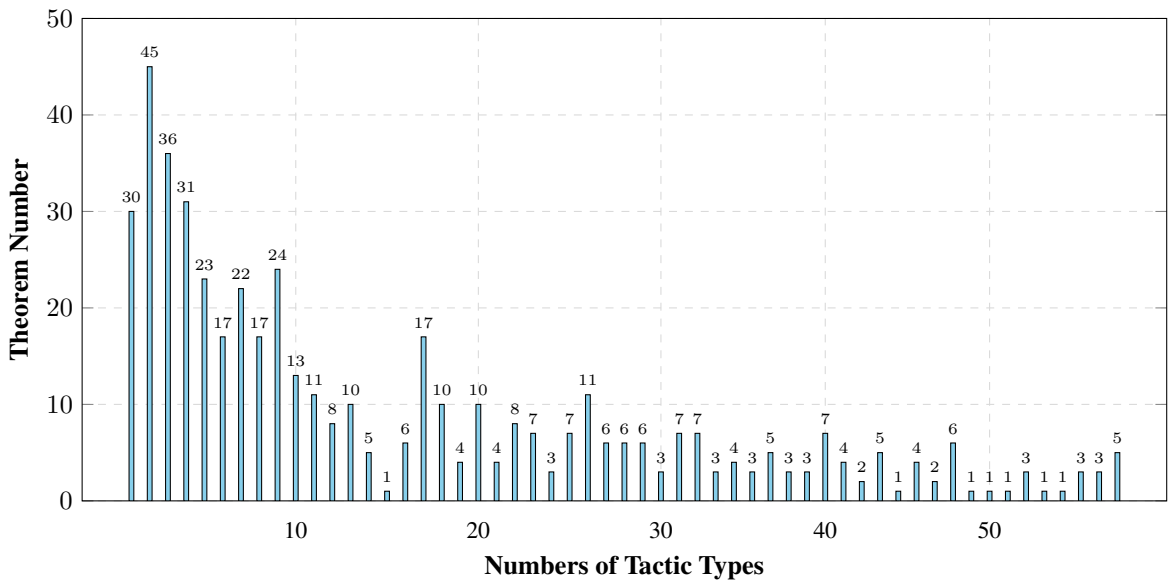
\begin{figure}[htp!]
\centering
\begin{tikzpicture}
    \begin{axis}[
        ybar,
        symbolic x coords={1,2,3,4,5,6,7,8,9,10,11,12,13,14,15,16,17,18,19,20,
                           21,22,23,24,25,26,27,28,29,30,31,32,33,34,35,37,38,39,
                           40,41,42,43,44,45,46,47,48,50,51,53,54,55,57,58,59 },
        xtick={10,20,30,40,50  }, % 每10显示一个刻度
        xtick pos=left,
        enlarge x limits=0.05,
        ymin=0, ymax=50,  
        xlabel={\textbf{Numbers of Tactic Types}},
        ylabel={\textbf{Theorem Number }},
        x tick label style={font=\footnotesize}, 
        ytick style={/pgf/number format/fixed},
        bar width=1.8pt,  % 缩窄柱子
        nodes near coords,  
        every node near coord/.append style={font=\tiny, black}, 
        grid=major,  
        major grid style={dashed, gray!30}, 
        xtick style={draw=none}, 
        width=16cm, height=8cm % 增大图表宽度
    ]
    \addplot[ybar, fill=SkyBlue] coordinates {
         (1,30)(2,45)(3,36)(4,31)(5,23)(6,17)(7,22)(8,17)(9,24)(10,13)
        (11,11)(12,8)(13,10)(14,5)(15,1)(16,6)(17,17)(18,10)(19,4)(20,10)
        (21,4)(22,8)(23,7)(24,3)(25,7)(26,11)(27,6)(28,6)(29,6)(30,3)
        (31,7)(32,7)(33,3)(34,4)(35,3)(37,5)(38,3)(39,3)(40,7)(41,4)
        (42,2)(43,5)(44,1)(45,4)(46,2)(47,6)(48,1)(50,1)
        
        (51,1)(53,3)
        (54,1)(55,1)(57,3)(58,3)(59,5) 
       
    };
    \end{axis}
\end{tikzpicture}
\caption{Theorem Distribution by Tactic Types of \textit{LeanComb} Benchmark $L^*$}
\label{fig:theorem_distribution}
\end{figure}

Figure \ref{fig:theorem_distribution} presents the distribution of theorem counts across different numbers of tactic types in the \textit{LeanComb} benchmark $L^*$. The chart shows that the majority of theorems are concentrated in the lower ranges of tactic types, with a peak occurring at tactic type 6. This suggests that simpler tactics are sufficient to prove most theorems, highlighting the prevalence of less complex theorem structures. However, as the number of tactic types increases, the theorem counts decrease significantly, indicating that highly complex theorems are relatively rare. This distribution implies that while simpler tactics are effective for many theorems, more sophisticated tactics are necessary for handling sparse but complex cases.

The distribution of tactic types is predominantly concentrated within the first ten types, with the highest theorem count occurring at six tactic types. Beyond this, the theorem count gradually decreases; however, hundreds to thousands of theorems remain even for tactic types ranging from 20 to 40. This demonstrates that the dataset contains many complex theorems, offering a significant challenge and utility for evaluating automated theorem-proving systems.

 \section{More Evaluation Results on \textit{LeanComb} Benchmark}
 
 \subsection{Evaluation on \textit{LeanComb} with Varying Time Limits and Candidate Tactic Numbers}

The results in Table \ref{tab_5} demonstrate the performance of several models on the \textit{LeanComb} test set under the constraints of a 300-second time limit and a maximum of 8 candidate tactics. The success rates across different models consistently show improvements when fine-tuned on \textit{LeanComb} and further enhanced with \textit{LeanComb}-Enhanced.

Gemma2-9B exhibits a modest increase, with a success rate of $13\%$ in the base setting, rising to $14\%$ with \textit{LeanComb} and further improving to $15\%$ with \textit{LeanComb}-Enhanced. Mathstral-3-8B demonstrates the most significant gains, increasing from $6\%$ in the base set to $11\%$ with \textit{LeanComb}, and reaching $21\%$ with \textit{LeanComb}-Enhanced, reflecting a substantial $+15\%$ improvement over the base model. Llama3-8B follows a similar trend, improving from $5\%$ to $10\%$ with \textit{LeanComb}, and to $18\%$ with \textit{LeanComb}-Enhanced. Mistral-7B, while starting with a slightly higher base performance of $8\%$, shows a smaller increase to $9\%$ with \textit{LeanComb}, but a more substantial jump to $20\%$ with the enhanced dataset.  

Under the same conditions, the model's proof success rate is generally lower than the proof success rate with a 600-second time limit and up to 16 candidate tactics, as shown in Table \ref{tab_4}, with a decrease range of $1\%$ to $8\%$.

\begin{table}[htp!]
\centering
\caption{Success rates on the \textit{LeanComb} test set with 300 seconds and 8 candidate tactics.}
\vspace{0.5em}
\label{tab_5}
\begin{tabular}{p{3cm}|>{\centering\arraybackslash}p{1.6cm} >{\centering\arraybackslash}p{1.8cm} >{\centering\arraybackslash}p{2.8cm}}
\toprule
\textbf{Model} & \textbf{Base} & \textbf{\textit{LeanComb}} & \textbf{\textit{LeanComb}-Enh.} \\ 
\midrule
Gemma2 - 9B   & $13\%$ & $14\%$ & $\textbf{15\%}$  \\ 
Mathstral3 - 8B & $6\%$ & $11\%$ & $\textbf{21\%}$ \\ 
Llama3 - 8B   & $5\%$ & $10\%$ & $\textbf{18\%}$ \\ 
Mistral - 7B    & $8\%$ & $9\%$ & $\textbf{20\%}$  \\ 
\bottomrule
\end{tabular}
\end{table}

These results indicate that all models benefit from the additional training data, with the \textit{LeanComb}-Enhanced dataset yielding the highest performance gains across the board. Notably, Mathstral3-8B and Llama3-8B exhibit the most pronounced improvements, suggesting that models with stronger baseline capabilities can leverage the enriched dataset more effectively. This underscores the value of \textit{LeanComb}-Enhanced in advancing automated theorem proving, particularly for combinatorial identities.

\subsection{Automated Proof Results of Our Models}

In this section, we analyze the proof process of the models. The table summarizes the average proof lengths across all models, revealing that the proof lengths range from $2.7$ to $3.9$. Notably, for both the Mathstral and Llama models, the enhanced versions consistently exhibit shorter proof lengths compared to their comb counterparts. This suggests that the models have learned to adopt more efficient and simpler tactics for generating proofs through extensive training on large datasets. For instance, as shown in the example for test\_087, Mathstral comb requires $11$ steps to complete the proof, whereas Llama3 enhanced accomplishes the proof in just four steps.

\begin{lstlisting}
theorem test_087_Mathstral_comb {n:ℕ} (hn : 0 < n):
  (ascPochhammer ℝ n).eval x = (x - 1 + n) * (ascPochhammer ℝ (n-1)).eval x := by
  unfold ascPochhammer
  cases n
  all_goals simp only [Polynomial.eval_one, CharP.cast_eq_zero, add_zero, mul_one]
  cases hn
  rename_i n
  cases n
  all_goals simp
  rename_i n
  rw [ascPochhammer_succ_right]
  simp only [Polynomial.eval_mul, Polynomial.eval_add, Polynomial.eval_X, Polynomial.eval_natCast]
  ring

theorem test_087_llama3_enhanced {n:ℕ} (hn : 0 < n):
  (ascPochhammer ℝ n).eval x = (x - 1 + n) * (ascPochhammer ℝ (n-1)).eval x := by
  have h1 : n = n - 1 + 1 := by
    rw [tsub_add_cancel_of_le]
    exact hn
  rw [h1, ascPochhammer_succ_right]
  simp only [Polynomial.eval_mul, Polynomial.eval_add, Polynomial.eval_X, Polynomial.eval_natCast,
    cast_add, cast_one, sub_add_add_cancel, add_tsub_cancel_right]
  rw [← mul_comm]
  
\end{lstlisting}

Furthermore, we observed that after training on the enhanced dataset, models tend to rely heavily on the theorems from the \textit{LeanComb} training set when generating proofs. However, despite this increased dependence on known theorems, the models occasionally introduce invalid or redundant steps in the proof process. For instance, in the following example, the step “lemma this” is an unnecessary operation; the proof remains valid even if this step is removed. Below, we present several interesting examples of automatically generated proofs:

\begin{lstlisting}
theorem test_062 (m n:ℕ)(hn:n≥2)(hm:m≥2) : 
Nat.choose (m+n-2) (m-1) = Nat.choose (m+n-3) (m-2)+Nat.choose (m+n-3) (m-1) := by
    have h1 : 2 ≤ m := by linarith
    have h2 : 2 ≤ n := by linarith
    have h3 : m - 1 + 1 = m := by omega
    have h4 : m - 2 + 2 = m := by
      rw [Nat.sub_add_cancel]
      exact h1
    have h5 : m + n - 2 = m + n - 3 + 1 := by omega
    rw [h5]
    rw [← h3, ← h4]
    simp only [Nat.add_one_sub_one, add_tsub_cancel_right, reduceSubDiff]
    rw [Nat.choose_succ_succ]

\end{lstlisting}

\begin{lstlisting}
theorem test_077 (n:ℕ) :
2 ^ n = ∑ k ∈ Finset.range (n + 1),(-1 : ℝ) ^ k * Nat.choose n k * 3 ^ (n - k) := by
  let f : ℝ → ℝ := fun x => x ^ n
  have h := add_pow (-1 : ℝ) 3 n
  have h1 : (-1 : ℝ) + 3 = 2 := by norm_num
  rw [h1] at h
  have h2 : ∑ k in Finset.range (n + 1), (-1 : ℝ) ^ k * 3 ^ (n - k) * (n.choose k) = ∑ k in Finset.range (n + 1), (-1 : ℝ) ^ k * (n.choose k) * 3 ^ (n - k) := by
   refine' sum_congr rfl fun k _ => _
   ring
  rw [h2] at h
  assumption
\end{lstlisting}

\begin{lstlisting}
theorem test_031 (n : ℕ) :
  ∑ k in Finset.range (n / 2 + 1), (-1 : ℝ) ^ k * Nat.choose (n - k ) k =(2 / Real.sqrt 3) *  Real.sin ((n + 1) * Real.pi / 3) := by
  obtain h1 := Idt_32 n (-1 : ℝ)
  have h2 :
    ∑ k in range (n / 2 + 1), ((n - k).choose k * (-1 : ℝ) ^ k) = ∑ k in range (n / 2 + 1), ((-1) : ℝ) ^ k * choose (n - k) k := by
      refine' sum_congr rfl fun k _ => _
      rw [mul_comm]
  rw [h2] at h1
  rw [h1]
  have : 1 + 4 * (-1 : ℝ) = -3 := by norm_num
  rw [this]
  exact_mod_cast complex_sqrt_neg n
\end{lstlisting}

\section {Experiments Details}
\label{AppendixF}

\subsection{Theorem Correction Details}

  \vspace{0.6em}
\newcommand{\tabincell}[2]{\begin{tabular}{@{}#1@{}}#2\end{tabular}}  
\begin{table*}[htp]
\centering
\resizebox{\textwidth}{46mm}{
\begin{tabular}{lcc}
\toprule
\textbf{Error Types} & \textbf{Error Example} & \textbf{Correction Example} \\
% \midrule 
% \textbf{Logical Errors} & 
% \tabincell{c}{$ErrorExample1(n : \mathbb{N} )$:\\
% $\sum_{k=1}^n nC_{n-1}{k-1} = n \sum_{k = 1}^{n-1}C_{n-1}$ := by\\
% $rw [sum\_Ico\_succ\_top]$\\
% --Incorrect Sub-goal: $1 \leq n$} 
% & Uncorrectable \\
\midrule 
\textbf{Incomplete Proofs} & 
\tabincell{c}{$ErrorExample_1(n : \mathbb{N} )$\\
(goal : the goal of sum\_mul\_congr) : \\
$\sum_{k = 0}^n k C_{2n+1} = 2n \sum_{k=0}^n C_{2n}^{k+1} \sum_{k=0}^n C_{2n}^k$ := by\\
$rw [range\_eq\_Ico]$ at goal\\
\textcolor{red}{assumption}} 
& 
\tabincell{c}{$CorrectedTheorem_1(n : \mathbb{N} )$\\
(goal : the goal of sum\_mul\_congr) : \\
$\sum_{k=0}^n k C_{2n+1} = 2n \sum_{k=0}^n C_{2n}^k + 1 \sum_{k=0}^n C_{2n}^k$ := by\\
$rw [range\_eq\_Ico]$ at goal\\
$rw [add\_mul]$ at goal\\
assumption}
\\
\midrule 
\textbf{Type Errors} & 
\tabincell{c}{$ErrorExample_2(m : \mathbb{N} )(hm : 0 < m) :$ \\
$2 + \textcolor{red}{-1} / (m + 1) = (2m + 1) / (m + 1)$} 
& 
\tabincell{c}{$Corrected\_Theorem_2(m : \mathbb{N} )(hm : 0 < m) :$ \\
$2 + (-1:R) / (m + 1) = (2m + 1) / (m + 1)$} 
\\
\midrule 
\textbf{Redundant Steps} & 
\tabincell{c}{$ErrorExample_3(n : \mathbb{N} ) :$ \\
$2n + 1 - n = n + 1$ := by\\
$rw [two\_mul]$\\
$rw [add\_assoc]$\\
$rw [add\_comm]$\\
simp\\
\textcolor{red}{$rw [two\_mul]$}}
& 
\tabincell{c}{$CorrectedTheorem_3(n : \mathbb{N} ) :$ \\
$2n + 1 - n = n + 1$ := by\\
$rw [two\_mul]$\\
$rw [add\_assoc]$\\
$rw [add\_comm]$\\
simp}
\\
\bottomrule
\end{tabular}}
\caption{The Error Types and Correction Process}
\label{table_Correction}
\end{table*} 
 \vspace{0.6em}
 
During the theorem generation process, we classify theorems and verify the correctness of candidate theorems. 
Incorrect theorems are grouped separately, and different correction methods are applied based on their types.

The standard MCTS with our fine-tuned Llama 3.1 
 corrects these theorems. We set the number of candidate tactics (i.e., visit counts) per node to $16$ and the number of simulations per node to $100$. It generates full-proof search trees through selection, expansion, and backpropagation, and complete-proof steps are engendered accordingly. The selection phase considers the average reward and exploration of nodes, using the UCB1 algorithm to select the optimal node ~\cite{auer2002finite}:

\begin{equation}
    UCB_1=\frac{W_{i}}{N_{i}}+C\times \sqrt{\frac{lnN_{p}}{N_{i}} }.  
\end{equation}

Detailed descriptions of the different types of errors and their respective correction methods are provided in the table \ref {table_Correction}. 

  \vspace{0.6em}
  
\textbf{Example1: Incomplete Proofs}
 \vspace{1.2em}
\begin{lstlisting}[caption={Type Error: unsolved goal $n - 1 < y$.}]
theorem congr_Ico_succ__2__73(n : ℕ) :
   ∑ k in Ico 1 (n + 1), k Nat.choose (n - 1) (k - 1) = ∑ l in
   Ico 0 n, (l + 1) Nat.choose (n - 1) l := by
  rw [sum_Ico_eq_sum_range]
  simp
  refine' sum_congr rfl fun y _ => _
  rw [add_mul]
  rW [choose_eq_zero_of_lt]
  rw [add_comm]
   
\end{lstlisting}\label{cl1}
 \vspace{0.6em}
  
\textbf{After Corrected:}
 \vspace{0.3em} 
\begin{lstlisting}
theorem congr_Ico_succ_2_73 (n : ℕ) :
   ∑ k in Ico 1 (n + 1), k Nat.choose (n - 1) (k - 1) = ∑ l 
   in Ico 0 n, (l + 1) Nat.choose (n - 1) l := by
  rw[sum_Ico_eq_sum_range]
  simp
  refine' sum_congr rfl fun x _ => _
  simp
  rw [add_comm]
  exact Or.inl rfl

\end{lstlisting}\label{cl2}

  \vspace{1.6em}
  
\noindent
\textbf{Example2: Type Errors}
 \vspace{0.3em}
\begin{lstlisting}[caption={Type Error : metavariables \ AddCommMonoid \ ?m.90801.}]
theorem sum_mul_add_distrib__0__36(n : ℕ)(h :  ∑ k in range (n + 1), (k + 1) Nat.choose n k = ∑ k in range (n + 1), (k*Nat.choose n k + 1*Nat.choose n k)) :
 (1 + y) Nat.choose n y = y * Nat.choose n y + Nat.choose n y := by
  refine' sum_congr rfl fun y _ => _
  simp [mul_assoc] at h
  rw [add_comm] at h
  assumption
  
\end{lstlisting}\label{cl3}
\vspace{1em}

\noindent
\textbf{After Corrected:}
 \vspace{0.3em}
\begin{lstlisting}
theorem sum_mul_add_distrib__0__36(n : ℕ)(h:  ∑ k in range (n + 1), (k + 1) Nat.choose n k = ∑ k in range (n + 1), (k Nat.choose n k + 1*Nat.choose n k)) :
   (1 + y)*Nat.choose n y = y*Nat.choose n y + Nat.choose n y := by
  rw [add_mul]
  rw [one_mul]
  rw [add_comm]
  
\end{lstlisting}
\vspace{1em}

\subsection{Details and Hyperparameters}

\begin{tcolorbox}[
  colback=SkyBlue!6, %  背景
  colframe= black, %  框线
    fonttitle=\bfseries\footnotesize,
  coltitle=titlecolor,
  rounded corners,
  boxrule=0.6mm
]
\textbf{Prompt Example:} \\

You are using Lean 4 for theorem proving. You are proving a theorem in Lean 4. Based on the current state of the theorem, provide the most reasonable proof tactic. Ensure your tactic is syntactically correct according to Lean 4's tactic syntax and effectively progresses the proof. \\

[Current State]: \\
n : $\mathbb{N}$ \\
$\vdash Nat.choose\ (2 * n)\ n\ \leq$ $\sum\  x\ \in\ range\ n,\ Nat.choose\ (2 * n)\ x\ +\ Nat.choose\ (2 * n)\ n$ \\

[Output Tactic]:

\begin{itemize}
\renewcommand\labelitemi{}
  \item rw [sum\_range\_add], \(-0.30987493962877327\)
  \item rw [sum\_range\_add\_choose] , \(-0.3457271686162461\)
  \item rw [sum\_range\_add\_choose\_add] , \(-0.3690287694334984\)
  \item rw [sum\_range\_add\_choose\_eq] , \(-0.44889247231185436\)
  \item rw [sum\_range\_add\_sum\_range] , \(-0.4555072601741323\)
  \item rw [sum\_range\_add] , \(-0.46009636194341713\)
  \item rw [sum\_range\_add\_choose] , \(-0.4931001846085895\)
  \item rw [← choose\_range\_choose] , \(-0.5118355770239511\)
  \item rw [← add\_assoc] , \(-0.5198788805936602\)
  \item rw [sum\_range\_choose\_add] , \(-0.5362029924396087\)
  \item rw [sum\_range\_add\_sum] , \(-0.5367179045623\)
  \item rw [sum\_range\_choose] , \(-0.5484384259002076\)
  \item rw [choose\_range\_choose] , \(-0.590293240112563\)
  \item rw [sum\_range\_succ] , \(-0.6008733134302828\)
  \item rw [two\_mul] , \(-0.6277757014509656\)
  \item simp , \(-1.4107935726642609\)
\end{itemize}

\end{tcolorbox}

In this experiment, we employed Llama-3.1-8B-Instruct as the candidate tactic model within our generator ATG4CI. This model is a transformer-based, autoregressive language model optimized for high performance. To enhance its alignment with human preferences in terms of utility and safety, we fine-tuned the model using Supervised Fine-Tuning (SFT)~\cite{wei2022finetunedlanguagemodelszeroshot} .
% and Reinforcement Learning from Human Feedback (RLHF)~\cite{ouyang2022traininglanguagemodelsfollow}.  
All training tasks were conducted on a cluster of six NVIDIA L40 GPUs (48GB each). For SFT, we employed Low-Rank Adaptation (LoRA)~\cite{hu2021loralowrankadaptationlarge}, leveraging bfloat16 mixed precision and DeepSpeed ZeRO Stage 0~\cite{jacobs2023deepspeedulyssesoptimizationsenabling} for memory optimization. LoRA was applied to all layers of the model. We used the AdamW optimizer~\cite{dettmers20228bitoquantization} with a batch size of 4. The learning rate was linearly warmed up from 0 to $5.0\times10^{-5}$ over the initial $5\% $ of training steps, followed by a cosine decay schedule for subsequent updates.  

We adopted a best-first search (BFS)~\cite{PearlBFS} tactic for evaluation. At each search step, the model generated 8 or 16 tactic candidates, deduplicated and ranked according to their log-likelihoods. Each candidate was subsequently verified using Lean 4, determining whether it should have been discarded or expanded into new proof states. The ranking of states is based on the cumulative log-likelihood of the tactics leading to the current state. The model is prompted using a structured (state, tactic) format, enabling the generation of tactics conditioned on the current proof state. 

% The prompt example is presented as follows:  

\end{document}